\documentclass[10pt,twocolumn,letterpaper]{article}

\usepackage[accsupp]{axessibility}  %

\usepackage{titling}
\usepackage{ifthen}
\newboolean{main}
\setboolean{main}{true}
\newboolean{suppmat}
\setboolean{suppmat}{true}

\usepackage[pagenumbers]{cvpr} %

\usepackage{graphicx}
\usepackage{amsmath}
\usepackage{amssymb}
\usepackage{booktabs}

\usepackage{appendix}
\usepackage{titletoc}

\def\cvprPaperID{*****} %
\def\confName{ICCV}
\def\confYear{2023}

\usepackage[dvipsnames]{xcolor}

\usepackage{graphicx}
\usepackage{amsmath}
\usepackage{amssymb}
\usepackage{booktabs}

\usepackage{cancel}

\usepackage{svg}

\usepackage{tikz}
\usetikzlibrary{positioning}
\usetikzlibrary{shapes.geometric}

\usepackage{breqn}
\usepackage{booktabs}
\usepackage{makecell}
\usepackage{multirow}
\usepackage{nicematrix}
\usepackage{soul}

\usepackage[noend,ruled,linesnumbered]{algorithm2e}
\SetKwInput{KwInput}{Input}                %
\SetKwInput{KwOutput}{Output}              %
\SetKwInput{KwInitialize}{Initialize}      %

\SetCommentSty{mycommfont}
\SetKwComment{Comment}{$\triangleright$\ }{}

\usepackage{amsthm}

\usepackage{booktabs}
\usepackage{multirow}
\usepackage{makecell}

\usepackage{array}
\newcolumntype{L}[1]{>{\raggedright\let\newline\\\arraybackslash\hspace{0pt}}m{#1}}
\newcolumntype{C}[1]{>{\centering\let\newline\\\arraybackslash\hspace{0pt}}m{#1}}
\newcolumntype{R}[1]{>{\raggedleft\let\newline\\\arraybackslash\hspace{0pt}}m{#1}}

\usepackage[pagebackref,breaklinks,colorlinks]{hyperref}
\hypersetup{
    colorlinks=true,
    allcolors=Blue,
    pdftitle={BANSAC: A dynamic BAyesian Network for adaptive SAmple Consensus},
    pdfauthor={V. Piedade and P. Miraldo},
    }

\usepackage[capitalize,nameinlink]{cleveref}
\crefdefaultlabelformat{#2\textbf{#1}#3}
\crefname{equation}{}{}
\creflabelformat{equation}{#2\textup{\bf #1}#3}
\crefname{equation}{Eq.}{Eqs.}
\Crefname{equation}{Equation}{Equations}
\crefname{figure}{Fig.}{Figs.}
\Crefname{figure}{Figure}{Figures}
\crefname{table}{Tab.}{Tabs.}
\Crefname{table}{Table}{Tables}
\crefname{section}{Sec.}{Secs.}
\Crefname{section}{Section}{Sections}
\crefname{problem}{Problem}{Problems}
\Crefname{problem}{Problem}{Problems}
\crefname{definition}{Definition}{Definitions}
\Crefname{definition}{Definition}{Definitions}
\crefname{lemma}{Lemma}{Lemmas}
\Crefname{lemma}{Lemma}{Lemmas}
\crefname{theorem}{Thm.}{Thms.}
\Crefname{theorem}{Theorem}{Theorems}
\crefname{remark}{Rmk.}{Rmks.}
\Crefname{remark}{Remark}{Remarks}
\crefname{enumi}{item}{items}
\Crefname{enumi}{Item}{Items}
\crefname{algocf}{Alg.}{Algs.}
\Crefname{algocf}{Algorithm}{Algorithms}
\crefname{assumption}{Asm.}{Asms.}
\Crefname{assumption}{Assumption}{Assumptions}
\crefname{ALC@unique}{line bla}{lines}
\Crefname{ALC@unique}{Line bla}{Lines}

\usepackage{enumitem}

\newlist{rquestions}{enumerate}{1}
\setlist[rquestions,1]{
    label={\bf RQ\arabic*:},
    ref=\arabic*, %
    labelwidth=!,
    align=left,
    itemindent=0pt,
    leftmargin=30pt,
    }
\creflabelformat{rquestionsi}{#2#1#3} %
\crefname{rquestionsi}{research question number}{research questions number} %
\Crefname{rquestionsi}{Research question number}{Research questions number}

\makeatletter
\def\maketag@@@#1{\hbox{\m@th\normalfont\normalsize#1}}
\makeatother

\begin{document}

\ifthenelse{\boolean{main}}
{

    \title{BANSAC: A dynamic BAyesian Network for adaptive SAmple Consensus}

    \author{
    Valter Piedade\\
    Instituto Superior T\'ecnico, Lisboa \\
    {\tt \href{mailto:valter.piedade@tecnico.ulisboa.pt}{valter.piedade@tecnico.ulisboa.pt}}
    \and
    Pedro Miraldo\\
    Mitsubishi Electric Research Labs \\
    {\tt \href{mailto:miraldo@merl.com}{miraldo@merl.com}}
    }

    \maketitle
    
    \begin{abstract}
    RANSAC-based algorithms are the standard techniques for robust estimation in computer vision. These algorithms are iterative and computationally expensive; they alternate between random sampling of data, computing hypotheses, and running inlier counting. Many authors tried different approaches to improve efficiency. One of the major improvements is having a guided sampling, letting the RANSAC cycle stop sooner. This paper presents a new adaptive sampling process for RANSAC. Previous methods either assume no prior information about the inlier/outlier classification of data points or use some previously computed scores in the sampling. In this paper, we derive a dynamic Bayesian network that updates individual data points' inlier scores while iterating RANSAC. At each iteration, we apply weighted sampling using the updated scores. Our method works with or without prior data point scorings. In addition, we use the updated inlier/outlier scoring for deriving a new stopping criterion for the RANSAC loop. We test our method in multiple real-world datasets for several applications and obtain state-of-the-art results. Our method outperforms the baselines in accuracy while needing less computational time. The code is available at \href{https://github.com/merlresearch/bansac}{https://github.com/merlresearch/bansac}.
    \end{abstract}

    \startcontents[corpstexte]
    \section{Introduction}
\label{sec:introduction}

\begin{figure}
    \resizebox{1\linewidth}{!}{%
        \setlength{\tabcolsep}{1pt}\begin{NiceTabular}{ccc}[code-before =%
        ]
        \makecell{\footnotesize Initialization\\
        \includegraphics[height=0.14\textheight]{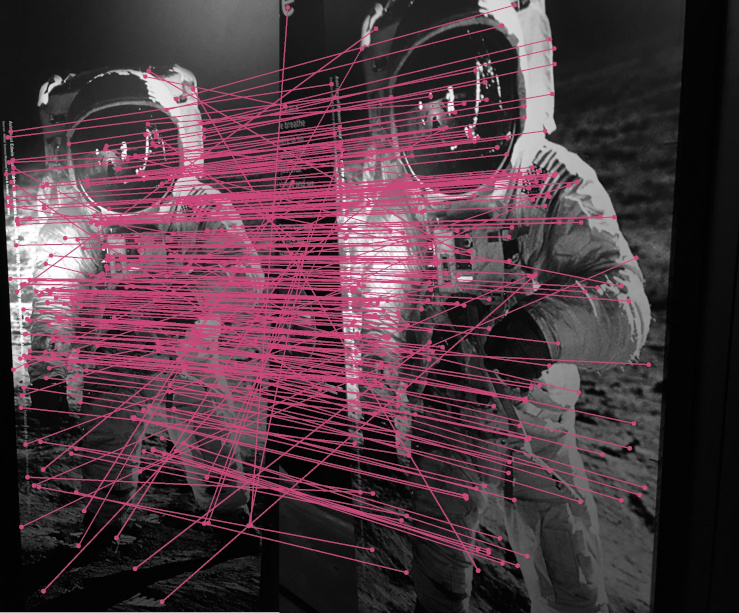}}
        & 
        \makecell{\footnotesize 10\textsuperscript{th} iteration\\
        \includegraphics[height=0.14\textheight]{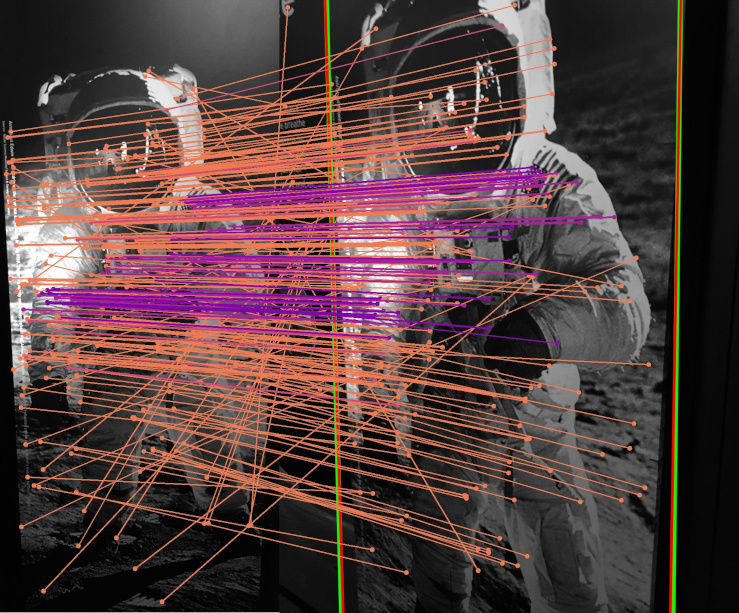}} & \multirow{2}{*}{\includegraphics[height=0.2\textheight]{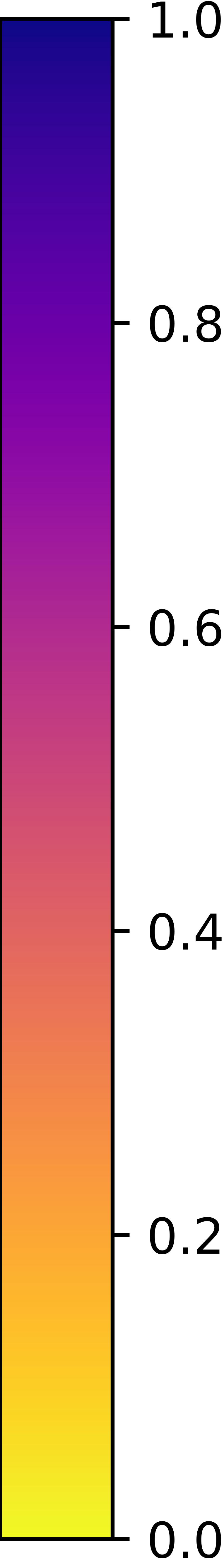}} \\
        \makecell{\footnotesize 100\textsuperscript{th} iteration\\
        \includegraphics[height=0.14\textheight]{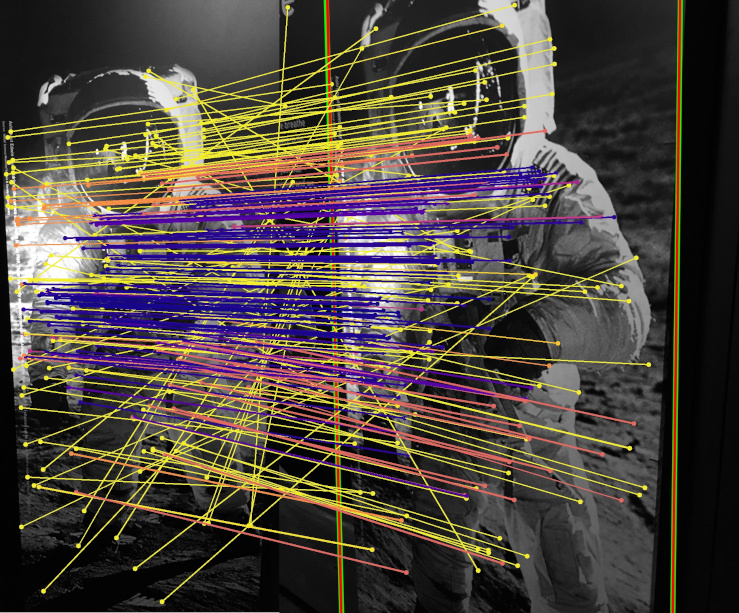}} & 
        \makecell{\footnotesize 1000\textsuperscript{th} iteration\\
        \includegraphics[height=0.14\textheight]{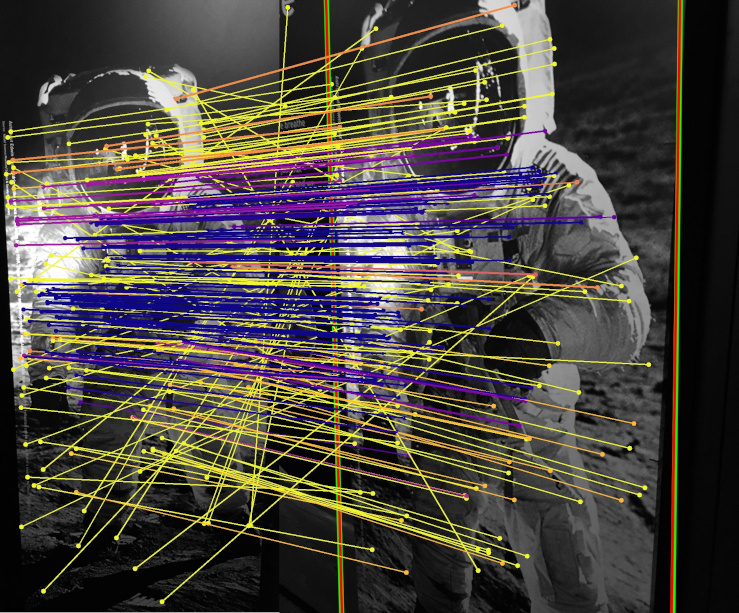}} &
    \end{NiceTabular}
    }
    \caption{\it We run BANSAC in a homography estimation problem. We took a pair from the HPatches dataset \cite{hpatches2017dataset} and show the updated inlier probabilities of data points (feature matches with color code at the right) over iterations. In the first row, from left to right, we show the probabilities at the start and iteration $10$. The second row shows iterations $100$ and $1000$. For visualization purposes, we show only $250$ randomly chosen matches.}
        \label{fig:teaser}
\end{figure}

Outliers are one of the primary causes of poor performance in computer vision. Robust estimators are essential since imaging sensors suffer from several types of noise and distortions. Removing outliers is one of the initial and more relevant steps in many computer vision tasks, such as relative pose estimation~\cite{mateus2020minimal,vakhitov2016accurate,miraldo2018minimal,pan2022camera,larsson2017making,cai2022ove6d}, camera localization~\cite{brachmann2018learning,williams2011automatic,sattler2011fast,sarlin2021back}, and mapping~\cite{schonberger2016pixelwise,schonberger2016structure,lindenberger2021pixel,orbslam2,ferrera2021ov,jancosek2011multi}. The gold-standard robust estimator is RANSAC (RANdom SAmple Consensus), introduced in \cite{fischler1981random}. RANSAC-based algorithms are iterative methods that, at each iteration: sample minimal sets, estimate a model, and run inlier counting. The output is the solution with the largest consensus.

The original RANSAC dates back to $1981$. Over the years, many authors changed the original loop to alleviate some of its limitations. All these alternatives focus on improving the sampling process, getting a better hypothesis, improving the stopping criteria, or changing the inlier counting. Most modifications add significant gains in computational efficiency. This paper focuses on improving the sampling efficiency even further.
The main question we want to tackle in this paper is: {\it Will changing the scoring weights over iterations help in sampling and defining the stopping criteria?} To answer this question, we propose BANSAC, a new sampling strategy for the RANSAC loop. \Cref{fig:teaser} illustrates our approach.

Previous methods such as~\cite{torr2002napsac,barath2019progressive,chum2005matching,ni2009groupsac,tordoff2005guided,brachmann2019neural} focus on exploiting scoring priors or considering some geometric relationships. However, the best-performing methods keep these scores fixed while running RANSAC. We focus on updating the scores online and using them for sampling minimal data. By modeling the problem with probabilities, we propose a new sampling strategy that uses a dynamic Bayesian network for updating the scores. These are the paper's main contributions:
\begin{itemize}
    \item[--] A novel adaptive sampling strategy that uses a dynamic Bayesian network to update data points' inlier scores. Our method does not need any prior information about the quality of the data, although it can use it;
    \item[--] A new simple and intuitive stopping criterion using the updated scores; and
    \item[--] Several experiments with multiple datasets show that our approach outperforms the best baselines in accuracy, being also more efficient.
\end{itemize}
We implemented BANSAC using C++, within the OpenCV USAC framework~\cite{opencv_library}. %

    \section{Related Work}

Over the years, RANSAC has been improving in a variety of areas. \cite{raguram2012usac} offers a single universal framework (USAC) that unites several improvements. Below we summarize some RANSAC improvements split into sampling and non-sampling strategies.

\subsection{Sampling strategies}
The original RANSAC~\cite{fischler1981random} assumes that every data point has the same likelihood of being an inlier. Several new sampling improvements have been proposed. We split these methods into: heuristic~\cite{torr2002napsac,barath2019progressive,chum2005matching,jo2015ransac,ni2009groupsac,cavalli2020handcrafted}, probabilistic~\cite{torr2000mlesac,tordoff2005guided,botterill2009new,fragoso2013evsac,mcilroy2010deterministic}, and learning approaches~\cite{brachmann2019neural,brachmann2017dsac,cavalli2022nefsac}.

\vspace{.25cm}\noindent
{\bf Heuristic-based strategies:}
Heuristic-based strategies take advantage of problem-specific characteristics to guide sampling. 
NAPSAC~\cite{torr2002napsac} assumes that points in high-density areas are more likely to be inliers. The algorithm chooses the first point randomly and completes the sample within a certain distance from the first. NAPSAC often leads to local or degenerate models for more complex problems. P-NAPSAC~\cite{barath2019progressive} improves some of NAPSAC issues by iteratively increasing the search space.
One of the most used sampling strategies is PROSAC~\cite{chum2005matching}. Using, \eg, similarity scores between point matches, PROSAC prioritizes the sampling of points with better scores. A drawback of this method is that it cannot be applied in general since it requires some previously computed score.
CS-RANSAC~\cite{jo2015ransac} argues that the matched features should neither be collinear nor adjacent to avoid degeneracies. CS-RANSAC defined the problem using a Constraint Satisfaction Problem (CSP) for homography matrix estimation. GroupSAC~\cite{ni2009groupsac} assumes that data can be split into groups according to their coordinates or based on the number of images observing the points.

\vspace{.25cm}\noindent
{\bf Probability-based strategies:}
Probabilistic-based sampling strategies such as~\cite{tordoff2005guided,fragoso2013evsac,meler2010betasac} focus on estimating prior probabilities for the data. These probabilities guide data selection during the sampling step. MLESAC~\cite{torr2000mlesac} improves the hypotheses verification process in fundamental matrix estimation. Guided-MLESAC~\cite{tordoff2005guided} further develops MLESAC by introducing a guided sampling modeled by two distinct distributions, one for matches and the other for mismatches. EVSAC~\cite{fragoso2013evsac} uses a Gamma and a Generalized Extreme Value distribution to model inliers and outliers, respectively. Although the problem differs from ours, \cite{mcilroy2010deterministic} derives an approach that uses a probability for modeling inlier/outlier classification over iterations utilizing multiple match hypotheses (a single feature on an image matches more than one feature on the second image).

The closest work to ours is BAYSAC~\cite{botterill2009new}, which updates the inlier probabilities iteratively, using it to guide the sampling. At each iteration, after choosing a minimal set of data points and computing the respective model hypothesis, the method updates the probability of the data points in the minimal set based on how good the hypothesis was. Although this method updates the inlier probability at each iteration, these updates are limited to the sampled points, which do not perform well without a good prior. In this paper, we propose a new approach in which all the data points' inlier probabilities are updated every iteration based on the inlier/outlier classifications.

\vspace{.25cm}\noindent
{\bf Learning-based strategies:}
There has been widespread use of neural networks in many areas of computer vision, RANSAC sampling being no exception. NG-RANSAC~\cite{brachmann2019neural} focuses on sampling by learning to estimate matching scores for the input correspondences for relative pose problems. Instead of scoring data points, DSAC~\cite{brachmann2017dsac} learns to score a set of previously computed hypotheses. NeFSAC~\cite{cavalli2022nefsac} predicts the probability that a minimal sample leads to an accurate solution, thus preventing the estimation of models using bad minimal samples.

Our method does not require training. However, pre-computed matching scores from learning-based solutions can be given as input to BANSAC.

\subsection{Non-sampling strategies}
Below we list some key works on improvements to RANSAC concerning the inlier threshold, inlier counting, local optimization, and stopping criteria.

\vspace{.25cm}\noindent
{\bf Inlier threshold:}
RANSAC uses the inlier ratio to select the best model (largest consensus). To compute the inlier ratio, a problem-dependent threshold is required. To avoid setting this parameter, MINPRAN~\cite{stewart1995minpran} proposes to model it using the model parameters. Alternatively, MAGSAC~\cite{barath2019magsac} and MAGSAC++~\cite{barath2020magsac++} reformulate the problem to use a weighted least squares fitting for model evaluation, using point scores as weights.

\vspace{.25cm}\noindent
{\bf Inlier counting:}
Inlier/outlier classification is computationally heavy since, in each iteration, all the data needs to be checked. Some authors have developed strategies to avoid scoring all the points every iteration, \cite{matas2004randomized,capel2005effective,matas2005randomized,chum2008optimal,barath2022learning}. Others check if the estimated models are valid, avoiding the scoring process for invalid models, such as \cite{chum2004epipolar,chum2005two,fan2022instability,barath2022learning,ivashechkin2021vsac}. To avoid scoring unnecessary data points, a bail-out test is proposed in~\cite{capel2005effective}. The scoring stops when the current model fails to have a higher inlier count than the best model. SPRT~\cite{matas2005randomized,chum2008optimal} estimates a likelihood ratio to decide if a model is good using the minimum possible amount of data.

\vspace{.25cm}\noindent
{\bf Local optimization:}
To improve accuracy, some authors added a new step called local optimization. LO-RANSAC in~\cite{chum2003locally} recomputes the model parameters when a new best model is found, using only the inliers. In~\cite{barath2018graph}, a method called Graph-Cut RANSAC is proposed. It takes advantage of the spatial coherence of the data to refine the estimated model by assuming that close neighbor points should have an equal classification in inlier/outlier. The data is arranged in a graph with edges between nearby points and is minimized by an energy cost function that penalizes neighbor points with different classifications.

\vspace{.25cm}\noindent
{\bf Stopping criteria:}
The stopping criterion checks if RANSAC found a good enough solution and can exit the loop.
The vanilla RANSAC in~\cite{fischler1981random} estimates how many iterations are needed until one all-inlier model hypothesis is selected based on the inlier ratio of the so-far best model. It stops when the current number of iterations is higher than the one needed for getting one all-inlier model. Instead of attempting to guarantee the best solution, \cite{nister2005preemptive} sets a real-time limit to get an estimate. PROSAC~\cite{chum2005matching} adds to the RANSAC criterion a condition to end when the probability of having a certain number of outliers in the current best set is lower than a predefined threshold. Finally, SPRT~\cite{matas2005randomized} terminates when the likelihood of missing a solution with a higher inlier set than the best solution found so far is below a certain threshold.

    \section{Background and Notations}

\begin{table}[t]
    \caption{\it Summary of some important notations used in this paper.}
    \label{tab:notations}
    \centering
    \resizebox{1\linewidth}{!}{\setlength{\tabcolsep}{2.5pt}\begin{NiceTabular}{@{}cp{6cm}@{}}[code-before =%
        \rectanglecolor{Gray!20}{2-1}{2-2}%
        \rectanglecolor{Gray!20}{4-1}{4-2}%
        \rectanglecolor{Gray!20}{6-1}{6-2}%
        \rectanglecolor{Gray!20}{8-1}{8-2}%
        ]
    \toprule
        \thead{Notation} & \thead{Description}  \\
        \midrule
        $\mathcal{X}^k \triangleq \{ x_1^k, ..., x_N^k \}$  & Inlier/outlier guesses per iteration. \\
        $\mathcal{X}_n \triangleq \{ x_n^0, ..., x_n^k \}$  & All inlier/outlier guesses for $\mathbf{x}_n$. \\
        $\mathcal{X} \triangleq \{\mathcal{X}^k, \mathcal{X}_n  \}$  & All data point inlier/outlier guesses. \\
        $\mathcal{C}_n \triangleq \{ c_n^1, ..., c_n^k \}$ & Set of inlier evidences per iteration. \\
        $\mathcal{C} \triangleq \{\mathcal{C}^k, \mathcal{C}_n  \}$  & All data point inlier/outlier evidences. \\
        $A^{0:m} \triangleq \{A^0, ..., A^m \}$ & Order subset. $A$ can be a variable or set. \\
        \makecell{$\begin{aligned} \mathcal{P}^k \triangleq \{ P(x_1^k \ | \ C_1^{1:k}), ..., \quad \\  P(x_N^k \ | \ C_N^{1:k}) \} \end{aligned}$} & \makecell[c]{Set of guesses for data points inlier/outlier \\ probabilities at $k$, given the evidence $c_{n}^{1:k}$.} \\
        \bottomrule
    \end{NiceTabular}
    }
\end{table}

RANSAC is an iterative method for solving a generic problem of type $f(\mathbf{x}, \theta) = 0$, where $\mathbf{x}$ is some data satisfying the model $\theta$. For simplicity, with a small abuse of notation, we call $\mathbf{x}$ a data point, because it can represent other types of features such as matches. The method iterates for a maximum of $K$ iterations while alternating between, 1)~sampling $\mathcal{S}^k \subset \mathcal{Q}$ data points, where $\mathcal{Q} \triangleq \{\mathbf{x}_1,...,\mathbf{x}_N\}$; 2) computing model hypothesis $\theta^k$; and 3) doing inlier counting, i.e., get $\mathcal{C}^k \triangleq \{ c_1^k, ..., c_N^k \}$, where $c_n^k$ is the inlier/outlier classification of $\mathbf{x}_n$ at iteration $k$. The output is the best-scored model and best inlier/outlier classification set, here denoted as $\{\theta^*, \mathcal{C}^*\}$.

    \section{BANSAC Method}\label{sec:sampling_strategy}

This paper focuses on deriving an efficient sampling of data points, i.e., getting $\mathcal{S}^k$. We note that getting $\theta^k$ from $\mathcal{S}^k$ is problem-dependent; BANSAC is independent of the problem. We take two simple assumptions:
\begin{enumerate}
    \item We assume that sampling data points with higher inlier scores gives a better model hypothesis; and \label{item:method_assuption_1}
    \item As we iterate through RANSAC, we get a better sense of whether $\mathbf{x}_n$ is an inlier or outlier. \label{item:method_assuption_2}
\end{enumerate}

An intuitive way of deriving a sampling technique with the above assumptions is to have changeable inlier/outlier scores of data points per iteration. Our method updates the scores based on inlier/outlier classifications from previous iterations. For modeling changeable scores over iterations, we define unknown variables, namely $x_n^k$ representing the best guess for inlier/outlier classification for a data point $\mathbf{x}_n$ at iteration $k$. For modeling the scores for each data point at each iteration, we use probabilities, i.e., $P(x_n^k)$, where $x_n^k$ can take the values of $\text{inlier}$ and $\text{outlier}$. Unfortunately, we do not have a direct way of measuring $P(x_n^k = \text{inlier})$. Instead, in this paper, we use inliers/outliers classifications obtained from previous iterations as evidence, i.e., for sampling we use $P(x_n^k = \text{inlier}\ |\ C_n^{1:k})$, where $C_n^k \triangleq c_n^k = \text{inlier/outlier}$. \Cref{tab:notations} lists some important notations we use in our derivations.

In the following subsection, we give a method overview.

\subsection{Algorithm outline}

\begin{algorithm}[t]
\small
    \caption{ {\it \small BANSAC algorithm outline}\newline
    \small {\bf Input} -- Data $\mathcal{Q}$, and (optional) pre-computed scores $\mathcal{P}^0$ \newline
    \small {\bf Output} -- Best model $\theta^*$, and $\mathcal{C}^*$
    }\label{alg:bansac}
    $k\gets 1$; \\
    \While{$k < K$}{
        $\mathcal{S}^k \gets \text{\underline{weighted\_sampling}}\left(\mathcal{Q}, \mathcal{P}^{k-1}\right)$\label{alg:sampling}\Comment*{\cref{sec:weighted_sampling}} 
        $\theta^k \gets \text{hypothesis}\left(\mathcal{S}^k\right)$;\label{alg:get_model}\\
        $\mathcal{C}^k \gets \text{model\_evaluation}\left(\mathcal{Q}, \theta^k \right)$;\label{alg:model_evaluation}\\
        $\theta^*, \mathcal{C}^* \gets \text{best\_model}\left(\mathcal{C}^k, \theta^k \right)$;\label{alg:best_model_update}\\
        $\mathcal{P}^k \gets \text{\underline{update\_probabilities}} \left( \mathcal{C}^k, \mathcal{X}^{0:k-1} \right)$\label{alg:update_probabilities}\Comment*{\cref{sec:probability_model}}
        {\bf if} $\text{\underline{stopping\_criteria}}\left(\mathcal{P}^k\right)$ \ {\bf break}\label{alg:stop}\Comment*{\cref{sec:stopping_criteria}} 
        $k \gets k+1$; \\
    }
\end{algorithm}

\Cref{alg:bansac} outlines the proposed method. The RANSAC-based loop starts with an optionally given $\mathcal{P}^0$, which can be obtained from matching scores (or other initial guesses), or set to a predefined value (i.e., not using previously pre-computed scores). At each iteration $k$, we generate a minimal set $\mathcal{S}^k\in\mathcal{Q}$ via weighted sampling, using $\mathcal{P}^{k-1}$ as weights (see notations in \cref{tab:notations}), which is shown in \cref{alg:sampling}. Next, we compute the hypothesis model $\theta^k$, run inlier counting, and update the best model if needed, as described in \cref{alg:get_model,alg:model_evaluation,alg:best_model_update}. \cref{alg:update_probabilities} updates the probabilities for the next iteration, i.e., computes $\mathcal{P}^k$. In addition to the new sampling and updating the probabilities, using $\mathcal{P}^k$, we derive a new stopping criterion for our RANSAC-based loop, which we use at \cref{alg:stop}.

\Cref{sec:probability_model} derives the probabilistic model, \cref{sec:weighted_sampling} describes the weighted sampling strategy, and \cref{sec:stopping_criteria} introduces the new stopping criterion.

\subsection{Probabilistic model and inference}\label{sec:probability_model}

We describe our method by modeling data points' inlier/outlier probability. 

\vspace{.15cm}\noindent
{\bf Probabilistic model:}
Since we want the data point probabilities to change over iterations, we use a dynamic Bayesian network (DBN) as our probabilistic model. A DBN is a probabilistic graphical model that uses variables (nodes representing states and observations) and their conditional dependencies (edges) in a directed acyclic graph (see \cite{russell2016artificial}). Starting with the variables, at each iteration $k$, we have the data points state and the inlier/outlier classifications (evidence) obtained so far (i.e., from $1$ to $k$). So, for iteration $k$, we have nodes $\mathcal{X}$ and $\mathcal{C}$ (see notations in \cref{tab:notations}).

For the graph edges, we have the following constraints:
\begin{enumerate}
    \item We want our sampling to be general. Then, for a certain iteration $k$, we assume that the inlier/outlier probabilities of different data points and the classifications are independent of each other. At each iteration $k$, the probability of $x_n^{k}$ is updated based on the $\mathbf{x}_n$'s previous probabilities and the previous inlier/outlier classifications (our evidence needs to constrain only the next probability estimate);
    \item The inlier/outlier evidence at each iteration, $c_n^k$, depends only on the model $\theta^k$, which depends on $\mathcal{S}^k$ and, by consequence, on $x_n^{k-1}$.
\end{enumerate}
Formally, for iteration $k$, we have the following constraints:
\begin{align}
    & x_n^k \perp \mathcal{X}  \setminus \mathcal{X}_n ,\ \mathcal{C} \setminus c_n^{k} \ \ | \ \ \mathcal{X}_n,\ c_n^{k} \label{eq:nodes_graph_1} \\
    & c_n^k \perp \mathcal{X}  \setminus x_n^{k-1} ,\ \mathcal{C} \ \ | \ \ x_n^{k-1}. \label{eq:nodes_graph_2}
\end{align}
The first important consequence of these conditionally independent constraints is that we have an independent DBN for each data point, $\mathbf{x}_n$, each with its own weights. Then, for iteration $k$, we define the DBN per data point as shown in \cref{fig:dynamic_bayes_net}.

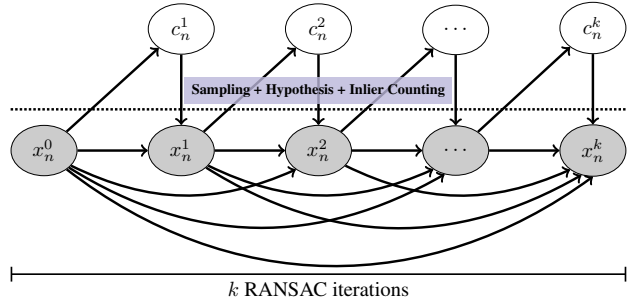
\begin{figure}[t]
    \centering%
    \scalebox{0.78}{%
        \begin{tikzpicture}

    \node[ellipse, draw, text = black, fill = black!20, minimum width=32pt, minimum height=24pt] (data0) at (0,0) {${x}_n^0$};
    \node[ellipse, draw, text = black, fill = black!20, right=1.2cm, at=(data0.east), minimum width=32pt, minimum height=24pt] (data1) {${x}_n^1$};
    \node[ellipse, draw, text = black, fill = white, above=1.2cm, at=(data1.north), minimum width=32pt, minimum height=24pt] (class1) {$c_n^1$};
    \draw[->, very thick] (class1.270) -- (data1.90);
    \draw[->, very thick] (data0.40) -- (class1.220);
    \draw[->, very thick] (data0.0) -- (data1.180);

    \node[ellipse, draw, text = black, fill = black!20, right=1.2cm, at=(data1.east), minimum width=32pt, minimum height=24pt] (data2) {${x}_n^2$};
    \node[ellipse, draw, text = black, fill = white, above=1.2cm, at=(data2.north), minimum width=32pt, minimum height=24pt] (class2) {$c_n^2$};
    \draw[->, very thick] (class2.270) -- (data2.90);
    \draw[->, very thick] (data1.40) -- (class2.220);
    \draw[->, very thick] (data1.0) -- (data2.180);
    
    \node[ellipse, draw, text = black, fill = black!20, right=1.2cm, at=(data2.east), minimum width=32pt, minimum height=24pt] (data3) {$\cdots$};
    \node[ellipse, draw, text = black, fill = white, above=1.2cm, at=(data3.north), minimum width=32pt, minimum height=24pt] (class3) {$\cdots$};
    \draw[->, very thick] (class3.270) -- (data3.90);
    \draw[->, very thick] (data2.40) -- (class3.220);
    \draw[->, very thick] (data2.0) -- (data3.180);

    \node[ellipse, draw, text = black, fill = black!20, right=1.2cm, at=(data3.east), minimum width=32pt, minimum height=24pt] (data4) {${x}_n^k$};
    \node[ellipse, draw, text = black, fill = white, above=1.2cm, at=(data4.north), minimum width=32pt, minimum height=24pt] (class4) {$c_n^k$};
    \draw[->, very thick] (class4.270) -- (data4.90);
    \draw[->, very thick] (data3.40) -- (class4.220);
    \draw[->, very thick] (data3.0) -- (data4.180);

    \draw[|-|, thick] ([yshift=-60pt]data0.west) -- node[below] {$k$ RANSAC iterations} ([yshift=-60pt]data4.east);

    \draw[densely dotted, very thick] ([yshift=20pt]data0.west) -- node[above,fill = Blue!25, yshift=2pt, opacity=.8,text opacity=1]  {\scriptsize\bf Sampling + Hypothesis + Inlier Counting} ([yshift=20pt]data4.east);

    \draw [draw = black, ->, very thick]
    (data0) edge [bend left=-25] (data2.220)
    (data1) edge [bend left=-25] (data3.220)
    (data2) edge [bend left=-25] (data4.220);

    \draw [draw = black, ->, very thick]
    (data0) edge [bend left=-31] (data3.245)
    (data1) edge [bend left=-31] (data4.245);

    \draw [draw = black, ->, very thick]
    (data0) edge [bend left=-37] (data4.270);
    
\end{tikzpicture}%
    }
    \caption{\it Dynamic Bayesian network proposed to model the data inlier/outlier probabilities for data point $\mathbf{x}_n$.}
    \label{fig:dynamic_bayes_net}
\end{figure}

\vspace{.15cm}\noindent
{\bf Markov assumptions:}
The DBN derived above has the unbounded problem of increasing exponentially with the number of iterations\footnote{Would need huge conditional probability tables (CPT) and the increase in computational cost.}. To solve this problem, we follow the typically used Markov assumptions. For simplicity, here we derive the first-order Markov assumption for our problem. The second and third-order assumptions are tested in \cref{sec:markov_assumptions_experiments}, and the derivations are provided in the supplementary materials; they are similar but slightly more intricated. In addition to the conditionally independent constraints derived above, we have
\begin{equation}
    x_n^{j} \perp x_n^{0:j-2}\ |\ x_{n}^{j-1}, c_{n}^{j}, \ \ \forall j,
    \label{eq:markov_assump_1}
\end{equation}
which means
\begin{equation}
    P(x_n^j \  |\ x_n^{0:j-1}, c_n^{j}) = P(x_n^j \  |\ x_n^{j-1}, c_n^{j}), \ \ \forall j.
    \label{eq:markov_assump_2}
\end{equation}

Now, by applying the chain rule of probabilities, we write the joint probability at iteration $k$ as 
\begin{equation}
    P(x_n^{0:k},c_n^{1:k}) = P(x_n^0) \prod_{j=1}^{k} \phi_j(x_n^{0:k}, c_n^{1:k}) ,
    \label{eq:joint_prob}
\end{equation} 
where
\begin{equation}
    \phi_j(x_n^{0:k}, c_n^{1:k}) = P(x_n^j \  |\ x_n^{j-1}, c_n^{j}) P(c_n^{j}\ |\ x_n^{j-1}).
    \label{eq:phi_markov_blanket}
\end{equation}

\vspace{.15cm}\noindent
{\bf Exact inference:}
In our sampling strategy, we use $P(x_n^k = \text{inlier}\ |\ C_n^{1:k})$, for all $n$.
This means that we are doing inference of $x_n^{k} = \text{inlier}$, with evidences $\mathcal{C}_n$, and hidden variables $\mathcal{X}_n \setminus x_n^{k}$. Given \cref{eq:phi_markov_blanket}, after some derivations, the exact inference is given by (see \cite[Sec.14.4]{russell2016artificial}):
\begin{equation}
    P(x_n^k=\text{inlier}\ |\ C_n^{1:k}) = \alpha \Phi(x_n^{k}=\text{inlier}, x_n^{0:k-1}, C_n^{1:k}),
    \label{eq:inference}
\end{equation}
where $\alpha$ is a normalization factor\footnote{From complementary rule, $\alpha$ is such that \newline $P(x_n^{k}=\text{inlier}, x_n^{0:k-1},C_n^{1:k}) + P(x_n^{k}=\text{outlier}, x_n^{0:k-1},C_n^{1:k}) = 1$}, and 
\begin{multline}
\Phi(x_n^k, x_n^{0:k-1},C_n^{1:k}) = \\
\sum_{x_n^{k-1}} \phi_k(x_n^{0:k}, c_n^{1:k})
\sum_{x_n^{k-2}} \phi_{k-1}(x_n^{0:k}, c_n^{1:k}) \\
\cdots
\sum_{x_n^1} \phi_2(x_n^{0:k}, c_n^{1:k})
\sum_{x_n^0} \phi_1(x_n^{0:k}, c_n^{1:k}).
\label{eq:Phi}
\end{multline}
Notice $x_n^k$ can take two values; it can be either inlier or outlier. This means that $\Phi(.)$ is a 2-dimension tuple; in \cref{eq:inference} we pick the case $x_n^k = \text{inlier}$. In addition, the summations for $x_n^j$ have two terms for all $j$.

A convenient result of \cref{eq:Phi} is that $\Phi(.)$ can be computed recursively as follows:
\begin{multline}
    \Phi(x_n^k, x_n^{0:k-1}, C_n^{1:k}) = \\ \sum_{x_n^{k-1}} \phi_k(x_n^{0:k}, c_n^{1:k}) \Phi(x_n^{k-1}, x_n^{0:k-2}, C_n^{1:k-1})
\end{multline}
for $k \geq 1$, and $ \Phi(x_n^0, -, -) = P(x_n^0)$. 
This means that at each iteration $k$, we can use the $\Phi(.)$ computed from the previous iteration, with no additional computational cost when increasing the number of iterations.
The pseudo-code for the probability update is in the supplementary materials.

We tried several alternatives for the Conditional Probability Tables (CPT) in \cref{eq:phi_markov_blanket}. In our experiments, for $P(c_n^k\ | \ x_n^{k-1})$, we use a Leaky ReLU, weighted using the inlier counting. For the CPT of $P(x_n^k\ | \ x_n^{k-1}, c_n^k)$, we get the probabilities empirically. Due to space limitations, we present more details about the CPTs in the supplementary material, including some experiments.

\subsection{Weighted sampling}\label{sec:weighted_sampling}
Weighted sampling aims at getting the minimal set $\mathcal{S}^k\subset\mathcal{Q}$ for computing hypothesis $\theta^k$. To increase the chances of only selecting inliers, we take the estimated probabilities $\mathcal{P}^{k-1}$ and create a weighted discrete distribution. Data points with a higher probability of being an inlier will have a higher chance of being sampled. 

In addition to directly using $\mathcal{P}^{k-1}$ for weighting the discrete distribution, we tested using various activation functions such as leaky ReLU, sigmoid, and tanh functions.
In the experiments, we directly use $\mathcal{P}^{k-1}$. Due to space constraints, we show results with other activation functions in the supplementary materials.

\subsection{Stopping criterion}\label{sec:stopping_criteria}
Besides using the probabilities $\mathcal{P}^{k}$ in sampling, we also exploit them in defining a stopping criterion. 
We know that $\mathcal{P}^{k}$ will influence the sampling, meaning that, after reaching a low probability threshold (which we denoted as $\tau$), we can say that a data point will not be considered for sampling.
Based on this idea, we derive a new, simple stopping criterion. At each iteration $k$, we add the following steps to \cref{alg:bansac}:
\begin{enumerate}
    \item In \cref{alg:best_model_update}, we keep the smallest number of outliers (best case scenario so far), denoted as $O^*$; and
    \item After updating the probabilities in \cref{alg:update_probabilities}, we compute the number of data points with $P(x_n^k = \text{inlier}\ | \ C_n^{1:k})$ lower than a predefined threshold $\tau$, which we denote here as $\widetilde{O}^k$.
\end{enumerate}
Our stopping criterion is triggered when $\widetilde{O}^k >= O^\ast$, which we check at each iteration in \cref{alg:stop}. This means that the current best model has a bigger or equal number of inliers than the set of $P(x_n^k = \text{inlier}\ | \ C_n^{1:k})$ being selected for sampling; it can be assumed we have only inliers in the sampling set. We exit the loop because there is a low chance of getting a better solution.

Our stopping criterion can be used alone or added to existing ones, such as the RANSAC~\cite{fischler1981random}, SPRT~\cite{matas2005randomized} or PROSAC~\cite{chum2005matching}. We do experiments on the combinations of these criteria in the supplementary materials.

\subsection{Degenerative configurations}

Degenerate configurations in minimal solvers can lead to poor RANSAC estimates. BANSAC is susceptible to such settings as other RANSAC-based methods. In a worst-case scenario, since the probability updates depend on the inlier ratio, after sampling degenerate configurations, BANSAC can create a bias towards degenerative sampling. However, it may take several iterations of consecutive degenerate configurations before that bias has some impact on the sampling. 
Moreover, since each data point always has a chance of being selected, even if it is minimal, BANSAC can reverse that bias as soon as new non-degenerative solutions are obtained.

Although possible, we highlight that we did not experience any issues with degenerative configurations during our experiments. In addition, we are using the OpenCV USAC framework that has built-in methods to handle many of those cases. When those degenerative configurations are detected, the probabilities are not updated.
    \section{Experiments}\label{sec:experiments}

We evaluate BANSAC in three computer vision problems: calibrated relative pose, uncalibrated relative pose, and homography estimation. We start our experiments with an ablation study to access the need for different Markov assumption orders in \cref{sec:markov_assumptions_experiments}. Next, we evaluate the efficiency of our method concerning a varying number of fixed iterations (no stopping criterion is used) and for a varying inlier ratio, in \cref{sec:number_iter,sec:outlier_ration}, respectively. \Cref{sec:results} offers results for 1) calibrated relative pose, 2) uncalibrated relative pose, 3) homography estimation, and 4) same as 1), 2) and 3) with the addition of local-optimization.

\vspace{.15cm}\noindent
{\bf Evaluation metrics:}
We use the mean Average Accuracy (mAA) with thresholds at $5$ and $10$ degrees for the calibrated and uncalibrated relative pose problems, and at $5$ and $10$ pixels for homography estimation. We borrow the evaluation scripts for rotation, translation, and homography error metrics from the ``RANSAC in 2020'' tutorial package\footnote{\href{https://github.com/ducha-aiki/ransac-tutorial-2020-data}{github.com/ducha-aiki/ransac-tutorial-2020-data} [\today]}.
Additionally, we show the average execution time.

\vspace{.25cm}\noindent
{\bf Methods:}
We utilize two variations of our method: with and without pre-computed scores. 
Without pre-computed scores, we refer to our method as BANSAC, and we use a combination of BANSAC (\cref{sec:stopping_criteria}) and SPRT~\cite{matas2005randomized} stopping criteria. When using pre-computed scores, we refer to our method as P-BANSAC, and we use a combination of BANSAC and PROSAC's stopping criteria~\cite{chum2005matching}. The stopping criteria for BANSAC and P-BANSAC are chosen for better accuracy (keeping a reasonable speed) and computational efficiency, respectively. In addition, BANSAC and P-BANSAC have some parameters that need to be set, namely the CPT values, which we kept fixed for all experiments.

As baselines, we use RANSAC~\cite{fischler1981random} and NAPSAC~\cite{torr2002napsac} when not using pre-computed scores, and P-NAPSAC~\cite{barath2019progressive} and PROSAC~\cite{chum2005matching} otherwise. Only the sampling and the stopping criterion vary between all methods. All the remaining RANSAC components are the same.

For all methods, at the end of the RANSAC cycle, the inliers of the best model are used to refine the final solution using a non-minimal solver, following a typical robust estimation pipeline. By default, no local optimization is run inside the RANSAC loop, unless we explicitly say so.

Experiments evaluating BANSAC using different stopping criteria and CPT parameters are present in the supplementary materials. BaySAC~\cite{botterill2009new} is not shown in the paper since its results do not differ much from RANSAC. In the supplementary materials, we evaluate BANSAC against RANSAC and BaySAC using synthetic data.

\vspace{.25cm}\noindent
{\bf Settings:}
For the calibrated relative pose problem, we use an error threshold of $1\mathrm{e}{-3}$ (normalized points), $1000$ maximum iterations, and a confidence of $0.999$. For the uncalibrated relative pose problem, we use an error threshold of $0.5$ pixel, $10000$ maximum iterations, and a confidence of $0.999$. For the homography estimation problem, we use an error threshold value of $1$ pixel, $1000$ maximum iterations, and a confidence of $0.999$. In all these problems, we set our proposed stopping criteria threshold $\tau$ to $0.01$ in BANSAC and $0.1$ in P-BANSAC (see \cref{sec:stopping_criteria}).

\vspace{.25cm}\noindent
{\bf Datasets:}
For the relative pose problems (essential and fundamental matrices estimation), we use the dataset ``CVPR IMW 2020 PhotoTourism challenge''~\cite{imageMatching}, which has 12 scenes with around 100K pairs and 2 sequences with around 5K pairs (inlier rates vary between $30$ and $60\%$, approximately). For the homography estimation problem, we use the EVD\footnote{\href{http://cmp.felk.cvut.cz/wbs/}{http://cmp.felk.cvut.cz/wbs/} [\today]} and HPatches~\cite{hpatches2017dataset}, with 7 and 145 pairs of images, respectively (we used the validation set since the test set does not provide ground-truth). Matches and pre-computed weights were obtained with RootSIFT features and nearest-neighbor matching. Dataloaders were borrowed from the ``RANSAC in 2020'' tutorial webpage.

For both relative pose problems, results were obtained by taking, for each scene, the first 4K pairs and repeating each trial 5 times to ensure we have replicable accuracy and computational time readings for different runs. In the homography matrix results, we use all the available pairs and repeat each trial 10 times. For the experiments with different Markov order assumptions, varying number of fixed iterations, and varying inlier ratio, we use the {\tt sacre\_coeur} sequence entirely (around 5K pairs). We also use this sequence to tune our BANSAC parameters (stopping criterion and probability update model).

\subsection{Different Markov assumption orders}
\label{sec:markov_assumptions_experiments}

\begin{table}[t]
    \centering
    \caption{\it Ablation study on the Markov assumption. We run BANSAC using the 1st, 2nd, and 3rd Markov assumption orders on a calibrated relative pose problem.}
    \label{table:ablation_markov_assumptions}
    \resizebox{1\linewidth}{!}{%
        \setlength{\tabcolsep}{7.5pt}\begin{NiceTabular}{@{}lccc}[code-before =%
        \rectanglecolor{Gray!20}{1-2}{7-4}%
        ]
        \toprule
        \multirow{2}{*}{\thead{Metrics}}&  \multicolumn{3}{c}{\thead{Markov assumptions}} \\ \cmidrule(lr){2-4}
        & \makecell{1st Order} & \makecell{2nd Order} & \makecell{3rd Order} \\ \midrule
            Rotation mAA $(5^\circ)$ \ $\uparrow$      & 0.836 & 0.822 & 0.808 \\
            Rotation mAA $(10^\circ)$ \ $\uparrow$     & 0.864 & 0.853 & 0.843 \\
            Translation mAA $(5^\circ)$ \ $\uparrow$   & 0.775 & 0.755 & 0.739 \\
            Translation mAA $(10^\circ)$ \ $\uparrow$  & 0.825 & 0.811 & 0.798 \\
            Avg. execution time $\left[ ms \right]$ \ $\downarrow$ & 13.9 & 12.2 & 11.8 \\
        \bottomrule
    \end{NiceTabular}
    }%
\end{table}

We start the experiments by running an ablation study on the Markov assumption described in \cref{sec:probability_model}. We use the 1st, 2nd, and 3rd Markov assumption orders (the 2nd and 3rd are derived in the supplementary materials) and run BANSAC for a calibrated relative pose problem.
Results are shown in \cref{table:ablation_markov_assumptions}. 

We observe that there are minor differences between the different Markov assumption orders. Rotation and translation errors are lower on the 1st-order assumption, and execution time is lower on the 3rd-order assumption. This occurs because probabilities converge more rapidly for higher orders, activating the stopping criterion faster. Prioritizing accuracy, in the following experiments, both BANSAC and P-BANSAC use the 1st-order Markov assumption in the probability modeling.

\subsection{Varying number of fixed iterations}\label{sec:number_iter}

\begin{figure}[t]%
    \includegraphics[height=0.165\textheight]{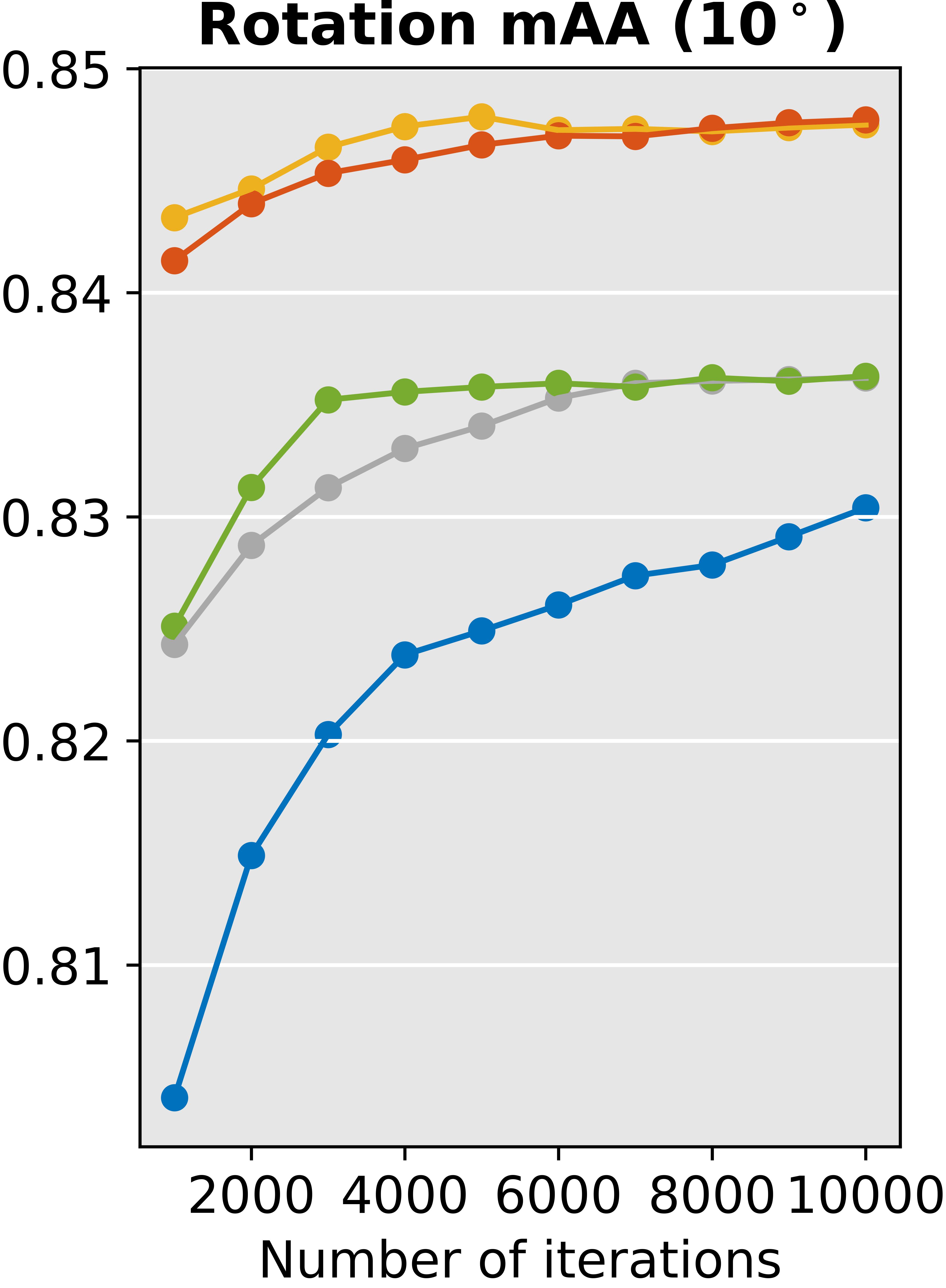}\hfill
    \includegraphics[height=0.165\textheight]{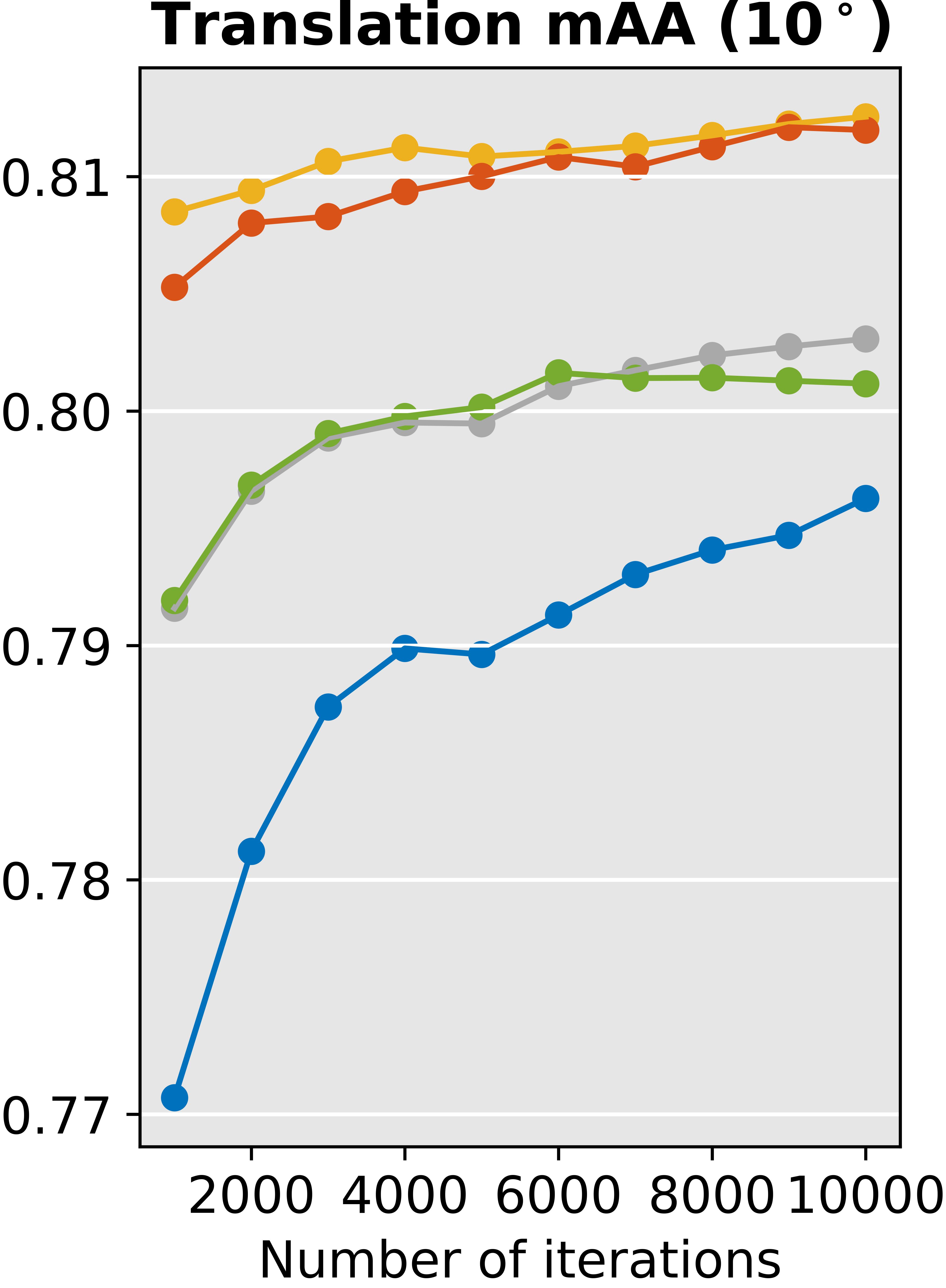}\hfill
    \includegraphics[height=0.165\textheight]{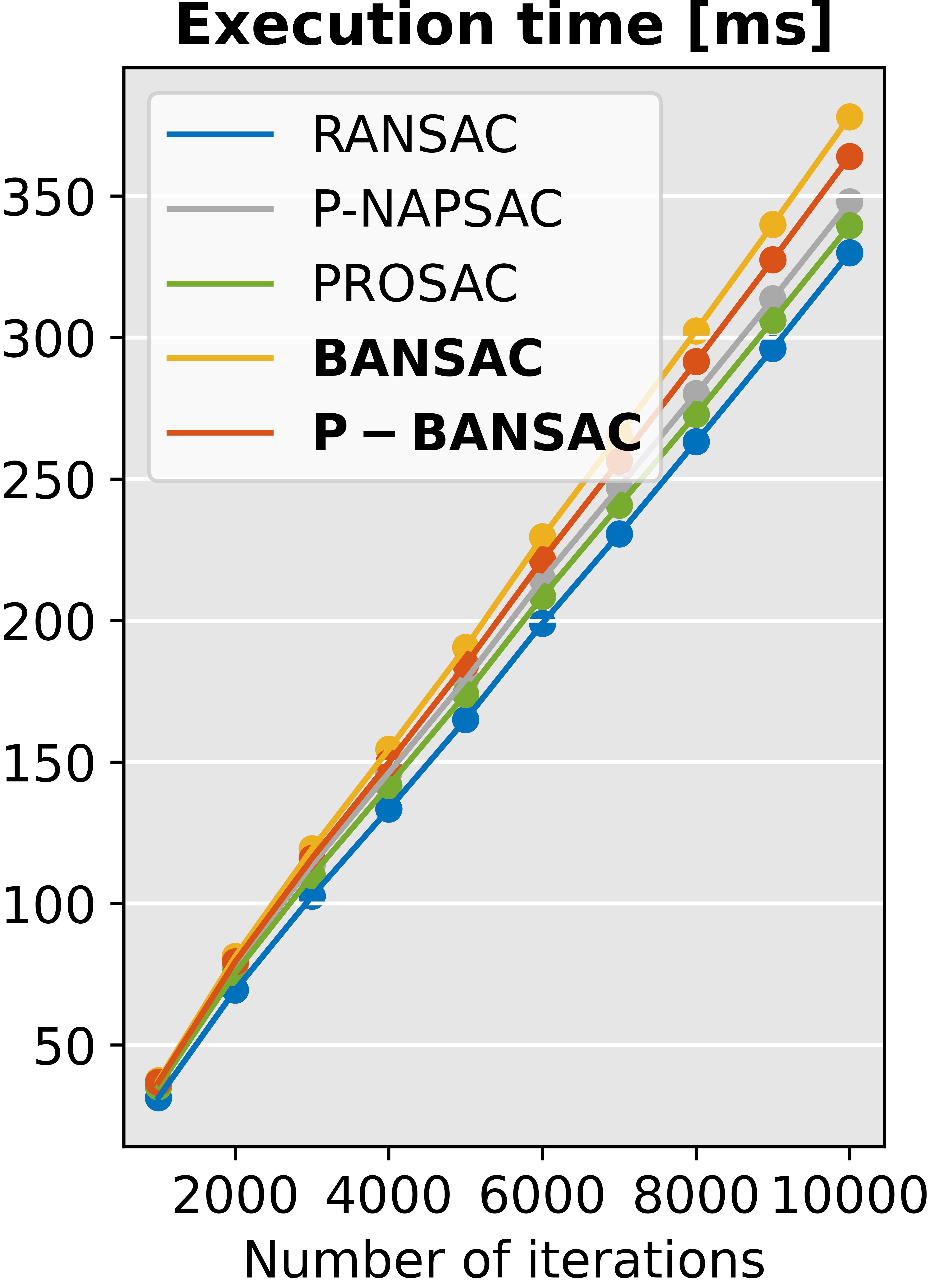}%
    \caption{\it Results for a fixed number of iterations, \ie, without stopping criterion. We vary the number of iterations between 1000 and 10000.}
    \label{fig:ablation_fixed_iters}
\end{figure}

In this experiment, we aim to evaluate the efficiency of the sampling process. We run experiments fixing the number of iterations for all the methods; no early stopping criterion is used. We vary the number of iterations between $1000$ and $10000$ on an uncalibrated relative pose problem and measure the rotation error, translation error, and execution time. The results obtained are shown in \cref{fig:ablation_fixed_iters}.%

As expected, overall, we verify that with an increasing number of iterations, the error decreases for all the methods. We observe that BANSAC and P-BANSAC have the lowest rotation and translation errors. Concerning execution time, although both our methods require additional steps to update the probabilities every iteration, we notice that the results are marginally the same.

\subsection{Varying inlier ratio}\label{sec:outlier_ration}

\begin{figure}[t]%
    \includegraphics[height=0.165\textheight]{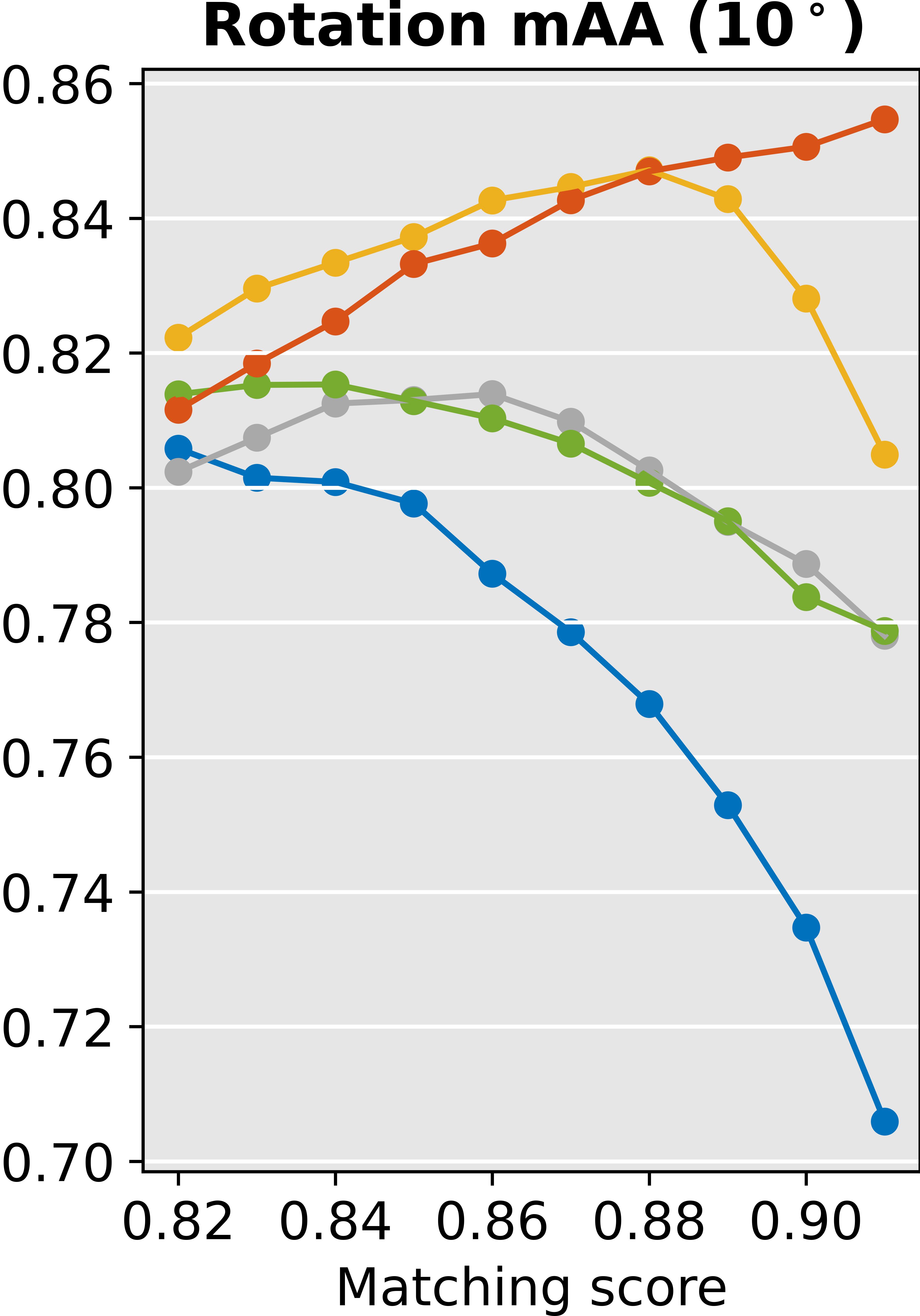}\hfill
    \includegraphics[height=0.165\textheight]{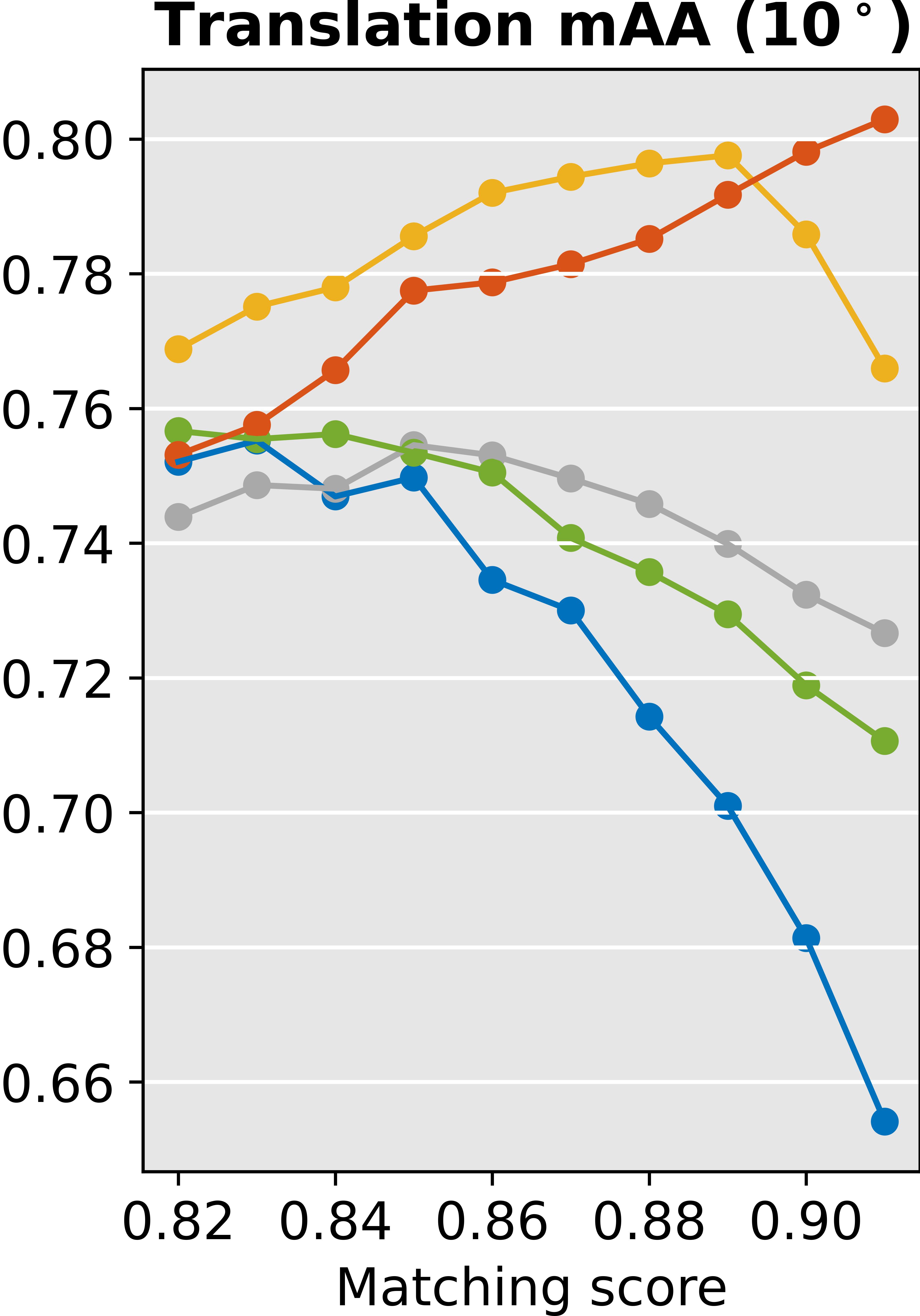}\hfill
    \includegraphics[height=0.165\textheight]{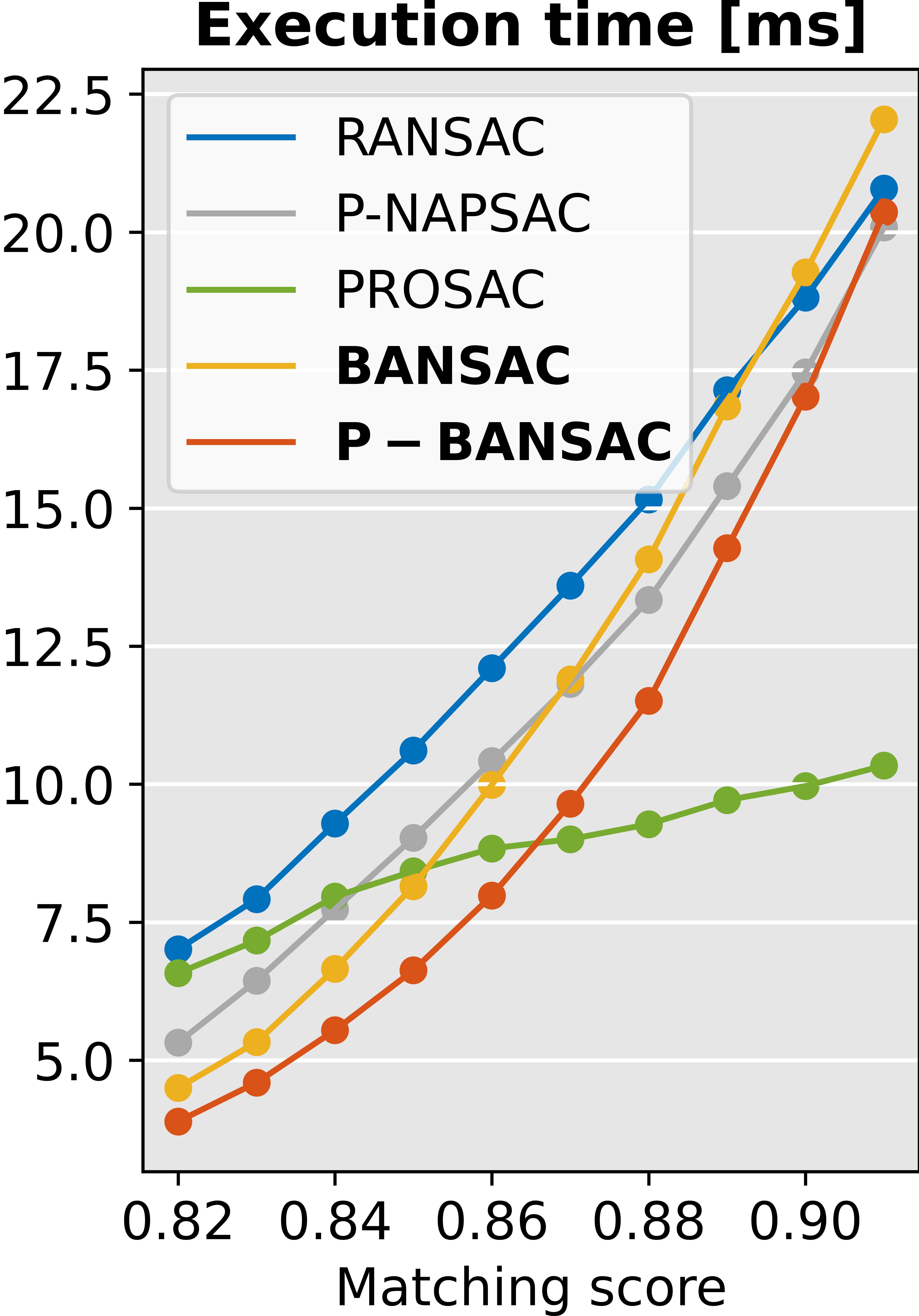}%
    \caption{\it Results varying the quality of the input data. As the matching score increases from 0.82 to 0.92, the inlier ratio decreases from 60\% to 30\%.}
    \label{fig:ablation_inlier_rates}
\end{figure}

The inlier ratio has a strong impact on RANSAC-based methods performance. To evaluate how each method performs for different inlier ratios, we vary the confidence threshold for filtering matches from $0.82$ to $0.92$, which gives us inlier rates ranging between around $60\%$ to $30\%$, respectively. Results for rotation and translation errors and execution time for an uncalibrated relative pose problem are shown in \cref{fig:ablation_inlier_rates}. %

In contrast to the baselines, we observe that the decrease in the inlier ratio (by filtering fewer matches) increases the accuracy for BANSAC and P-BANSAC.
Concerning execution time, it increases in all methods similarly, except in PROSAC where it grows less.

\subsection{Results}\label{sec:results}
Next, we present results for three computer vision problems with and without a local-optimization step.

\begin{table*}[t]
    \caption{\it Experimental results for the calibrated relative pose, uncalibrated relative pose, and homography estimation. We compare BANSAC and P-BANSAC with RANSAC~\cite{fischler1981random}, NAPSAC~\cite{torr2002napsac}, P-NAPSAC~\cite{barath2019progressive}, and PROSAC~\cite{chum2005matching}. We show results with and without the local-optimization step of LO-RANSAC~\cite{chum2003locally}.}
    \label{tab:results_E_F_H_LO}
    \resizebox{1\linewidth}{!}{%
        \setlength{\tabcolsep}{1.5pt}\begin{NiceTabular}{@{}l@{\hskip .5pt} c c c c c c c c c c c c }[code-before =%
        \rectanglecolor{gray!20}{2-6}{29-7}%
        \rectanglecolor{gray!20}{2-12}{29-13}%
        ]
            \toprule
            \Block{2-1}{\thead{Metrics}} & \multicolumn{6}{c}{\bf Without Local Optimization} & \multicolumn{6}{c}{\bf With  Local Optimization} \\ \cmidrule(l){2-7} \cmidrule(l){8-13}
            & \thead{RANSAC} & \thead{NAPSAC} & \thead{P-NAPSAC} & \thead{PROSAC} & \thead{\bf BANSAC} & \thead{\bf P-BANSAC} & \thead{RANSAC} & \thead{NAPSAC} & \thead{P-NAPSAC} & \thead{PROSAC} & \thead{\bf BANSAC} & \thead{\bf P-BANSAC} \\\midrule\\[-8pt]
            & \multicolumn{12}{c}{\it Calibrated Relative Pose Estimation (essential matrix estimation)} \\[2pt]%
                    {Rotation mAA $(5^\circ)$ \ $\uparrow$}                  & 0.568 & 0.158 & 0.551 & 0.569 & \textbf{0.610} & \ul{0.603}  & 0.569 & 0.216 & 0.557 & 0.570 & \textbf{0.611} & \ul{0.604} \\
                    {Rotation mAA $(10^\circ)$ \ $\uparrow$}                 & 0.645 & 0.226 & 0.641 & 0.653 & \textbf{0.680} & \ul{0.675}  & 0.645 & 0.292 & 0.645 & 0.655 & \textbf{0.680} & \ul{0.675} \\
                    {Translation mAA $(5^\circ)$ \ $\uparrow$}               & 0.422 & 0.0810 & 0.402 & 0.417 & \textbf{0.460} & \ul{0.454} & 0.423 & 0.114 & 0.409 & 0.419 & \textbf{0.461} & \ul{0.454} \\
                    {Translation mAA $(10^\circ)$ \ $\uparrow$}              & 0.532 & 0.137 & 0.514 & 0.527 & \textbf{0.566} & \ul{0.559}  & 0.532 & 0.176 & 0.520 & 0.528 & \textbf{0.566} & \ul{0.560} \\
                    {Avg. execution time $\left[ ms \right]$ \ $\downarrow$} & 25.5  & 40.1  & 20.9  & 21.5  & \ul{15.6}  & \textbf{15.2}   & 27.6  & 42.9  & 26.6  & 22.6  & \ul{18.0}  & \textbf{17.4}  \\
                    \midrule\\[-12pt]%
            & \multicolumn{12}{c}{\it Uncalibrated Relative Pose Estimation (fundamental matrix estimation)} \\[2pt]%
                    {Rotation mAA $(5^\circ)$ \ $\uparrow$}                  & 0.467 & 0.206 & 0.460 & 0.464 & \textbf{0.500} & \ul{0.478}  & 0.514 & 0.499 & \ul{0.517} & 0.511 & \textbf{0.526} & 0.501 \\
                    {Rotation mAA $(10^\circ)$ \ $\uparrow$}                 & 0.559 & 0.308 & 0.557 & 0.560 & \textbf{0.589} & \ul{0.571}  & 0.595 & 0.572 & \ul{0.600} & 0.595 & \textbf{0.610} & 0.589 \\
                    {Translation mAA $(5^\circ)$ \ $\uparrow$}               & 0.267 & 0.0780 & 0.260 & 0.264 & \textbf{0.292} & \ul{0.274} & 0.308 & 0.300 & \ul{0.309} & 0.307 & \textbf{0.317} & 0.294 \\
                    {Translation mAA $(10^\circ)$ \ $\uparrow$}              & 0.353 & 0.129 & 0.345 & 0.349 & \textbf{0.380} & \ul{0.360}  & 0.394 & 0.381 & \ul{0.396} & 0.392 & \textbf{0.405} & 0.381 \\
                    {Avg. execution time $\left[ ms \right]$ \ $\downarrow$} & 14.4  & 26.3  & 12.2  & 12.5  & \ul{9.80 } & \textbf{7.88}   & 15.4  & 18.4  & 13.6  & 13.5  & \ul{11.4}  & \textbf{8.87}  \\
                    \midrule\\[-12pt]%
            & \multicolumn{12}{c}{\it Homography Estimation} \\[2pt]%
                    {Homography mAA $(5 \ px)$ \ $\uparrow$}    & 0.422 & 0.152 & 0.158 & 0.210 & \ul{0.443} & \textbf{0.446} & 0.513 & 0.498 & 0.488 & 0.333 & \textbf{0.542} & \ul{0.517} \\
                    {Homography mAA $(10 \ px)$ \ $\uparrow$}  & 0.552 & 0.208 & 0.235 & 0.291 & \ul{0.569} & \textbf{0.573} & 0.647 & 0.617 & 0.617 & 0.426 & \textbf{0.672} & \ul{0.650} \\
                    {Avg. execution time $\left[ ms \right]$ \ $\downarrow$} &  \textbf{2.30} &  \ul{2.78} &  2.96 &  3.09 & 4.04  & 3.07  &  \ul{3.24} &  5.94 &  6.43 &  \textbf{2.89} &  4.09 &  4.33 \\
            \bottomrule%
        \end{NiceTabular}%
     }%
\end{table*}

\vspace{0.25cm}\noindent
{\bf Calibrated relative pose:}
Results for the calibrated relative pose (essential matrix estimation) are shown in \cref{tab:results_E_F_H_LO}. They show that BANSAC is the most accurate method, followed by P-BANSAC across all the scenes. In execution time, P-BANSAC is the fastest method, followed by BANSAC, indicating that the pre-computed scores help BANSAC on exiting the RANSAC loop earlier.

\vspace{0.25cm}\noindent
{\bf Uncalibrated relative pose:}
Results for the uncalibrated relative pose (fundamental matrix estimation) are shown in \cref{tab:results_E_F_H_LO}. Similarly to the previous results, we observe that BANSAC is consistently the most accurate method, and P-BANSAC is the second most accurate in most scenes. In runtime, P-BANSAC is the fastest and BANSAC the second fastest. \Cref{fig:prob_update_fundamental} shows the probabilities values after $10$, $100$, $1000$, and $10000$ iterations in an image pair from the {\tt sacre\_coeur} sequence, using BANSAC.

\begin{figure}[t]
    \resizebox{1\linewidth}{!}{%
        \setlength{\tabcolsep}{1pt}\begin{NiceTabular}{cc}[code-before =%
        ]
        \makecell{\large 10\textsuperscript{th} iteration\\
        \includegraphics[width=0.7\linewidth]{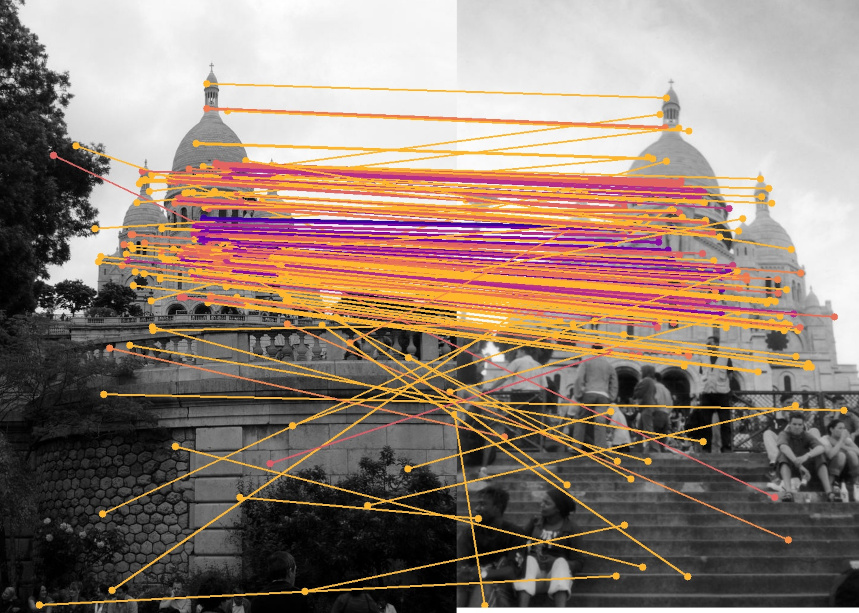}}
        & 
        \makecell{\large 100\textsuperscript{th} iteration\\
        \includegraphics[width=0.7\linewidth]{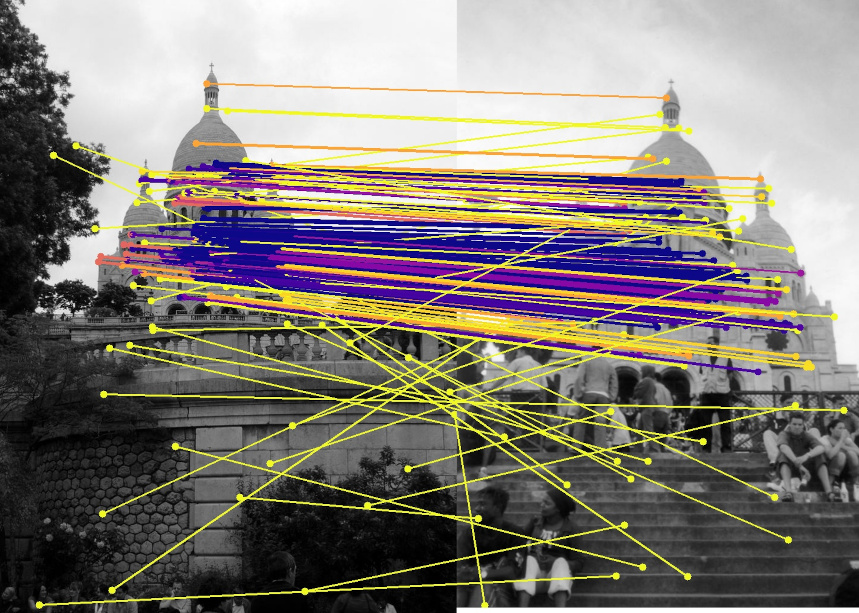}} \\
        
        \makecell{\large 1000\textsuperscript{th} iteration\\
        \includegraphics[width=0.7\linewidth]{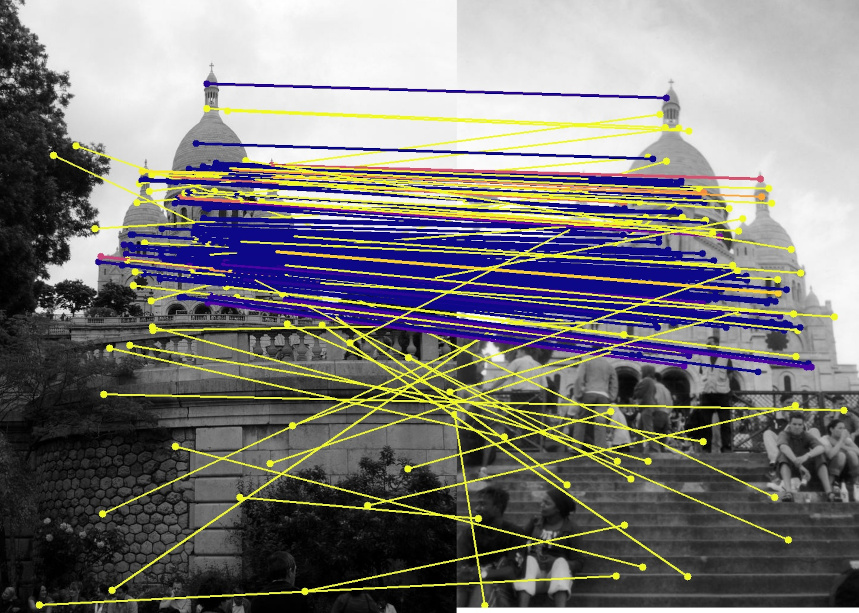}}
        &
        \makecell{\large 10000\textsuperscript{th} iteration\\
        \includegraphics[width=0.7\linewidth]{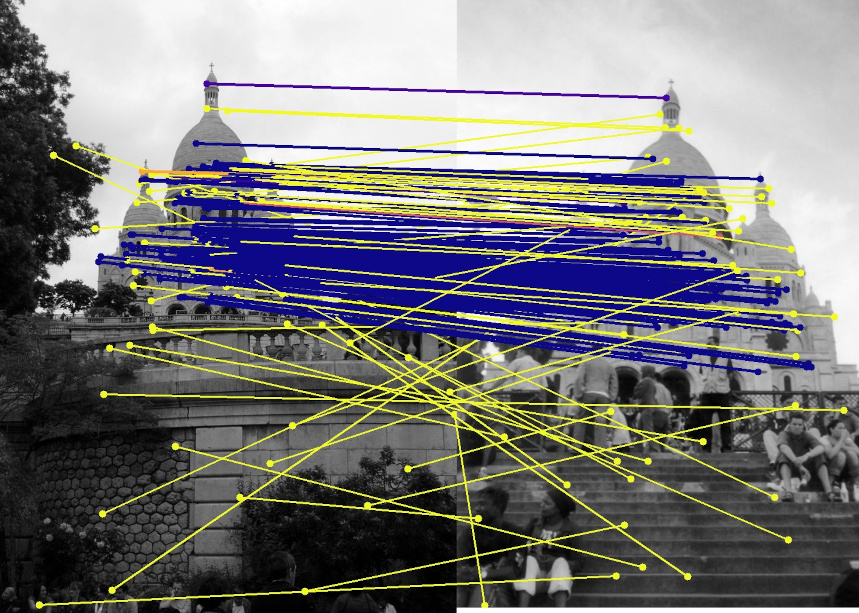}}\\

        \multicolumn{2}{c}{\includegraphics[height=0.033\textheight]{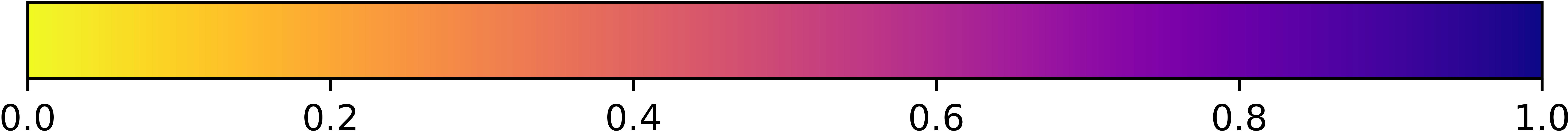}} \\
    \end{NiceTabular}
    }
    \caption{\it Data inlier probabilities over iterations for a fundamental matrix estimation problem using BANSAC (color code at the right). Image pair is from the PhotoTourism {\tt sacre\_coeur} scene. For visualization purposes, we show only $250$ randomly chosen matches.}
        \label{fig:prob_update_fundamental}
\end{figure}

\vspace{0.25cm}\noindent
{\bf Homography estimation:}
Results for homography estimation are shown in \cref{tab:results_E_F_H_LO}.
In this experiment, P-BANSAC is the best in accuracy, with BANSAC being second best.
In runtime, RANSAC is the fastest. We note that BANSAC requires an additional loop over all data points per iteration for updating scores (see the discussion in \cref{sec:conclusion}) when compared with RANSAC. This extra computational effort is visible when BANSAC does not exit the loop sufficiently earlier than the baselines, as shown in \cref{fig:ablation_fixed_iters}.
\Cref{fig:teaser} shows the initial probabilities of some randomly chosen matches (all started at 0.5) and the updated probabilities after $10$, $100$, and $1000$ iterations in an image pair from the HPatches dataset using BANSAC, demonstrating the probability updates over iterations. The ground truth and the estimated homography are marked in green and red, respectively.

\vspace{0.25cm}\noindent
{\bf Local optimization:}
We add the local-optimization (LO) step in \cite{chum2003locally} to all methods and repeat the previous experiments. Results are present in \cref{tab:results_E_F_H_LO}. We observe that LO improves the accuracy for every method, with an increase in execution time. We also note that the improvement in accuracy is significant for some of the baselines, \eg, NAPSAC and P-NAPSAC. Overall, BANSAC and P-BANSAC continue to outperform the baselines by some margin.

    \section{Conclusion}\label{sec:conclusion}

This paper proposes BANSAC, a new sampling strategy for RANSAC using dynamic Bayesian networks. The method performs weighted sampling using probabilities for scoring data points. These probabilities are adaptively updated every iteration based on the successive inlier/outlier classifications. Additionally, we propose a stopping criterion using the estimated probabilities.

We present results on challenging real-world datasets showing that the proposed algorithm can learn the data inlier probability and that these probabilities can guide the sampling efficiently; the updates to RANSAC bring improvements in accuracy and execution time.

We note that BANSAC updates the scoring weights depending on the quality of the hypothesis, which is obtained from inlier counting. This means that we need an extra loop cycle every iteration for updating the scores. Although we do not observe a significant increase in computational cost compared to the best baselines, there is room for improvement. In future work, we plan to include a more efficient hypothesis prediction for incorporating the probability update in the inlier counting loop.

\section*{Acknowledgments}

Valter Piedade was supported by the National Centre for Research and Development under the Smart Growth Operational Programme as part of project {\tt POIR.01.01.01-00-0102/20} and by the LARSyS$-$FCT Project {\tt UIDB/50009/2020}. We thank all the reviewers and ACs for their valuable feedback.

    {\small
    \bibliographystyle{ieee_fullname}
    \bibliography{egbib}
    }
}
{}

\ifthenelse{\boolean{suppmat}}
{
    \title{BANSAC: A dynamic BAyesian Network for adaptive SAmple Consensus\\
    \large\textsc{(Supplementary Materials)}
    }
    
    \author{
    Valter Piedade\\
    Instituto Superior T\'ecnico, Lisboa \\
    {\tt \href{mailto:valter.piedade@tecnico.ulisboa.pt}{valter.piedade@tecnico.ulisboa.pt}}
    \and
    Pedro Miraldo\\
    Mitsubishi Electric Research Labs \\
    {\tt \href{mailto:miraldo@merl.com}{miraldo@merl.com}}
    }
    
    \maketitle

    \startcontents[annexes]
    \appendix
    \renewcommand\thefigure{\thesection.\arabic{figure}}    
    \renewcommand\thetable{\thesection.\arabic{table}}    

    {\it \noindent
    These supplementary materials present new quantitative experiments (\cref{supp:experiments}) and some additional derivations and pseudo-code (\cref{sec:other_markov_assumptions}). 
    }
    \section*{Contents}
    \printcontents[annexes]{l}{0}{\setcounter{tocdepth}{3}}

\section{Additional Experiments}\label{supp:experiments}

This section provides additional experiments with real-world and synthetic data. \Cref{supp:essential,supp:fundamental} show results with each scene from the PhotoTourism dataset for the essential and fundamental matrices estimation.
\Cref{supp:synthetic} offers results for curve and circle fitting problems using synthetic data. \Cref{supp:abaltion} contains ablation studies on the conditional probability tables (CPTs), sampling weights, and stopping criteria.

For both of the relative pose problem experiments (\cref{supp:essential,supp:fundamental}), we use the following scenes from the PhotoTourism dataset, with a matching score cutoff of 0.85: 0) {\tt brandenburg\_gate} with $43\%$ inliers; 1) {\tt palace\_of\_westminster} with $32\%$ inliers; 2) {\tt westminster\_abbey} with $49\%$ inliers; 3) {\tt taj\_mahal} with $57\%$ inliers; 4) {\tt prague\_old\_town\_square} with $32\%$ inliers; and 5) {\tt st\_peters\_square} with $46\%$ inliers; 6) {\tt buckingham\_palace} with $45\%$ inliers; 7) {\tt colosseum\_exterior} with $36\%$ inliers; 8) {\tt grand\_place\_brussels} with $31\%$ inliers; 9) {\tt notre\_dame\_front\_facade} with $46\%$ inliers; 10) {\tt pantheon\_exterior} with $62\%$ inliers; 11) {\tt temple\_nara\_japan} with $60\%$ inliers; 12) {\tt trevi\_fountain} with $33\%$ inliers; and 13) {\tt sacre\_coeur} with $51\%$ inliers. As in the main document, we use 4K pairs for each scene and repeated each trial 5 times.

All experiments presented in this document and on the main paper were performed on an Intel(R) Core(TM) i7-7820X CPU @ 3.60GHz processor.

\subsection{Calibrated relative pose}\label{supp:essential}

\begin{table*}[t]
    \caption{\it Experimental results for the calibrated and uncalibrated relative pose estimation problems for each scene in the PhotoTourism dataset.}
    \label{tab:photoTourism_results}
    \resizebox{1\linewidth}{!}{%
        \setlength{\tabcolsep}{8.7pt}\begin{NiceTabular}{@{}l@{\hskip 7pt}| c c c c c c c c c c c c}[code-before =%
        \rectanglecolor{Gray!20}{2-6}{50-7}%
        \rectanglecolor{Gray!20}{2-12}{50-13}%
        ]
            \toprule
            \Block{2-1}{\thead{Seq.}} & \multicolumn{6}{c}{\thead{Calibrated Relative Pose Estimation (essential matrix estimation)}} & \multicolumn{6}{c}{\thead{Uncalibrated Relative Pose Estimation (fundamental matrix estimation)}} \\ \cmidrule(lr){2-7}\cmidrule(lr){8-13}
            & \thead{RANSAC} & \thead{NAPSAC} & \thead{P-NAPSAC} & \thead{PROSAC} & \thead{BANSAC} & \thead{P-BANSAC} & \thead{RANSAC} & \thead{NAPSAC} & \thead{P-NAPSAC} & \thead{PROSAC} & \thead{BANSAC} & \thead{P-BANSAC}\\ \midrule\\[-8pt]
               & \multicolumn{12}{c}{\it Rotation mAA $(10^\circ)$ \ $\uparrow$}\\[6pt]%
            0  & 0.711 & 0.245 & 0.740 & 0.754 & \ul{0.773} & \textbf{0.775} & 0.574 & 0.260 & 0.585 & \ul{0.593} & \textbf{0.608} & \ul{0.593} \\
            1  & 0.555 & 0.218 & 0.604 & 0.612 & \ul{0.624} & \textbf{0.625} & 0.482 & 0.253 & 0.504 & 0.514 & \textbf{0.546} & \ul{0.537} \\ 
            2  & 0.714 & 0.417 & 0.709 & 0.714 & \textbf{0.719} & \ul{0.717} & 0.686 & 0.455 & 0.684 & 0.686 & \textbf{0.693} & \ul{0.689} \\ 
            3  & \textbf{0.866} & 0.259 & 0.797 & 0.820 & \textbf{0.866} & 0.848 & \ul{0.863} & 0.567 & 0.859 & 0.857 & \textbf{0.883} & \textbf{}0.863 \\ 
            4  & \ul{0.322} & 0.142 & 0.290 & 0.302 &\textbf{ 0.331} & 0.311 & \ul{0.269} & 0.111 & 0.267 & 0.246 & \textbf{0.280} & 0.264 \\ 
            5  & 0.759 & 0.251 & 0.745 & 0.772 & \ul{0.803} & \textbf{0.804} & 0.628 & 0.336 & 0.617 & 0.617 & \textbf{0.661} & \ul{0.642} \\ 
            6  & 0.684 & 0.216 & 0.658 & 0.693 & \textbf{0.730} & \ul{0.724} & 0.569 & 0.275 & 0.566 & 0.571 & \textbf{0.605} & \ul{0.576} \\ 
            7  & 0.434 & 0.136 & 0.448 & 0.449 & \ul{0.467} & \textbf{0.468} & 0.374 & 0.191 & 0.375 & 0.377 & \textbf{0.409} & \ul{0.394} \\ 
            8  & 0.357 & 0.133 & 0.359 & 0.368 & \textbf{0.380} & \ul{0.379} & 0.301 & 0.186 & 0.295 & 0.300 & \textbf{0.317} & \ul{0.306} \\ 
            9  & 0.669 & 0.271 & 0.698 & 0.716 & \ul{0.730} & \textbf{0.731} & 0.582 & 0.258 & 0.593 & 0.599 & \textbf{0.635} & \ul{0.625} \\
            10 & 0.762 & 0.204 & 0.718 & 0.707 & \textbf{0.786} & \ul{0.784} & \ul{0.467} & 0.306 & 0.420 & 0.434 & \textbf{0.480} & 0.438 \\ 
            11 & \ul{0.829} & 0.255 & 0.797 & 0.815 & \textbf{0.838} & 0.816 & \ul{0.762} & 0.484 & 0.746 & 0.744 & \textbf{0.783} & 0.728 \\ 
            12 & 0.532 & 0.217 & 0.568 & 0.579 & \textbf{0.605} & \ul{0.598} & 0.458 & 0.212 & 0.468 & 0.471 & \textbf{0.501} & \ul{0.490} \\ 
            13 & 0.827 & 0.201 & 0.836 & 0.844 & \textbf{0.867} & \ul{0.862} & 0.804 & 0.418 & 0.819 & 0.819 & \textbf{0.846} & \ul{0.839} \\ 
            \midrule \\[-8pt]%
               & \multicolumn{12}{c}{\it Translation mAA $(10^\circ)$ \ $\uparrow$} \\[6pt]  
            0  & 0.581 & 0.150  & 0.599 & 0.613 & \ul{0.643} & \textbf{0.647} & 0.363 & 0.0970 & 0.360 & \ul{0.377} & \textbf{0.394} & 0.376 \\
            1  & 0.504 & 0.162  & 0.548 & 0.558 & \ul{0.565} & \textbf{0.567} & 0.418 & 0.183  & 0.436 & 0.446 & \textbf{0.480} & \ul{0.466} \\ 
            2  & 0.494 & 0.184  & 0.480 & 0.486 & \textbf{0.506} & \ul{0.501} & 0.377 & 0.121  & 0.373 & 0.377 & \textbf{0.390} & \ul{0.384} \\ 
            3  & \ul{0.641} & 0.116  & 0.544 & 0.570 & \textbf{0.649} & 0.626 & \ul{0.610} & 0.265  & 0.597 & 0.596 & \textbf{0.635} & 0.609 \\ 
            4  & \ul{0.282} & 0.0970 & 0.244 & 0.253 & \textbf{0.292} & 0.267 & \ul{0.176} & 0.0390 & 0.167 & 0.153 & \textbf{0.192} & 0.167 \\ 
            5  & 0.601 & 0.141  & 0.570 & 0.601 & \textbf{0.642} & \ul{0.635} & 0.351 & 0.121  & 0.328 & 0.331 & \textbf{0.379} & \ul{0.357} \\ 
            6  & 0.622 & 0.178  & 0.590 & 0.627 & \textbf{0.661} & \ul{0.651} & 0.274 & 0.110  & 0.281 & 0.287 & \textbf{0.311} & \ul{0.294} \\ 
            7  & 0.403 & 0.107  & 0.409 & 0.409 & \ul{0.432} & \textbf{0.434} & 0.257 & 0.100  & 0.252 & 0.259 & \textbf{0.296} & \ul{0.279} \\ 
            8  & 0.274 & 0.0850 & 0.267 & 0.277 & \textbf{0.297} & \ul{0.296} & 0.140 & 0.0590 & 0.137 & 0.140 & \textbf{0.151} & \ul{0.141} \\ 
            9  & 0.592 & 0.209  & 0.614 & 0.631 & \ul{0.655} & \textbf{0.662} & 0.413 & 0.150  & 0.416 & 0.430 & \textbf{0.461} & \ul{0.456} \\
            10 & 0.611 & 0.114  & 0.540 & 0.524 & \textbf{0.620} & \textbf{0.620} & \ul{0.213} & 0.0770 & 0.169 & 0.180 & \textbf{0.214} & 0.185 \\ 
            11 & \ul{0.617} & 0.0830 & 0.549 & 0.563 & \textbf{0.630} & 0.600 & \ul{0.378} & 0.108  & 0.338 & 0.332 & \textbf{0.389} & 0.323 \\ 
            12 & 0.417 & 0.122  & 0.446 & 0.453 & \textbf{0.491} & \ul{0.486} & 0.217 & 0.0460 & 0.215 & 0.215 & \textbf{0.247} & \ul{0.235} \\
            13 & 0.798 & 0.173  & 0.792 & 0.800 & \textbf{0.837} & \ul{0.832} & 0.757 & 0.329  & 0.761 & 0.759 & \textbf{0.792} & \ul{0.777} \\ 
            \midrule \\[-8pt]%
               & \multicolumn{12}{c}{\it Avg. execution time $\left[ ms \right]$ \ $\downarrow$} \\[6pt]
            0  & 26.6 & 40.3 & 19.9 & 20.5 & \textbf{17.3} & \ul{17.4} & 15.4 & 29.4 & \ul{10.9} & 11.8 & 12.6 & \textbf{9.75} \\
            1  & 34.9 & 43.3 & 29.1 & 31.1 & \ul{21.7} & \textbf{20.6} & 21.9 & 30.2 & 18.8 & 20.1 & \ul{14.6} & \textbf{12.3} \\ 
            2  & 17.9 & 35.8 & 15.4 & 15.2 & \textbf{10.1} & \ul{10.7} & 10.1 & 25.9 & 12.6 & 9.67 & \ul{5.84} & \textbf{5.43} \\ 
            3  & 13.0 & 37.1 & 9.48 & 9.49 & \ul{9.02} & \textbf{8.02} & 7.90 & 25.5 & 5.76 & \ul{5.60} & 6.72 & \textbf{4.49} \\ 
            4  & 33.5 & 42.4 & 28.8 & 28.8 & \ul{17.2} & \textbf{15.9} & 20.7 & 28.8 & 18.4 & 18.1 & \ul{11.6} & \textbf{8.82} \\ 
            5  & 22.1 & 37.9 & 17.3 & 16.8 & \ul{15.8} & \textbf{15.6} & 12.7 & 24.6 & 9.39 & \ul{9.29} & 11.5 & \textbf{8.92} \\ 
            6  & 28.0 & 39.9 & 21.7 & 23.5 & \ul{18.1} & \textbf{17.8} & 14.4 & 27.1 & \ul{9.17} & 12.1 & 10.6 & \textbf{8.29} \\ 
            7  & 32.4 & 42.2 & 28.1 & 29.9 & \ul{17.1} & \textbf{16.5} & 19.4 & 27.5 & 18.5 & 18.8 & \ul{9.86} & \textbf{8.81} \\ 
            8  & 37.6 & 43.2 & 32.7 & 34.4 & \ul{19.7} & \textbf{19.4} & 22.3 & 26.3 & 20.1 & 21.2 & \ul{11.7} & \textbf{10.3} \\ 
            9  & 26.6 & 40.1 & 21.2 & 22.0 & \textbf{16.4} & \textbf{16.4} & 14.7 & 28.3 & 12.1 & 12.4 & \ul{10.1} & \textbf{8.12} \\
            10 & 13.4 & 38.2 & \textbf{8.96} & 10.7 & \ul{9.41} & 9.73 & 5.67 & 12.3 & \ul{4.43} & 4.72 & 5.08 & \textbf{3.34} \\ 
            11 & 13.8 & 36.9 & \ul{8.90} & 8.98 & 9.33 & \textbf{8.85} & 5.16 & 16.3 & \ul{3.71} & 3.45 & 4.51 & \textbf{2.61} \\ 
            12 & 37.1 & 43.2 & 32.0 & 32.6 & \textbf{22.7} & \ul{23.2} & 21.4 & 30.8 & 19.2 & 19.4 & \ul{14.4} & \textbf{13.0} \\
            13 & 21.4 & 41.3 & 17.1 & 16.8 & \textbf{14.7} & \ul{14.8} & 11.3 & 35.0 & 9.74 & 9.15 & \ul{8.61} & \textbf{7.12} \\ 
            \bottomrule%
        \end{NiceTabular}%
}
\end{table*}

This subsection presents additional results for the calibrated relative pose estimation problem, comparing BANSAC and P-BANSAC against the baselines (RANSAC, NAPSAC, P-NAPSAC, and PROSAC). As estimation parameters, we use an error threshold of $1\mathrm{e}{-3}$ (normalized points), $1000$ maximum iterations, and a confidence of $0.999$, and set the BANSAC stopping criteria threshold $\tau$ to $0.01$ in BANSAC and $0.1$ in P-BANSAC (same parameters as in the main paper's results). Results are shown in \cref{tab:photoTourism_results}.

We observe that, in accuracy, BANSAC and P-BANSAC are the best methods overall. In execution time, P-BANSAC is the best, with BANSAC second best in most scenes.

\subsection{Uncalibrated relative pose}\label{supp:fundamental}

This subsection presents further results for the uncalibrated relative pose estimation problem, using the same baselines as in the previous subsection. As estimation parameters, we use an error threshold of $0.5$, $10000$ maximum iterations, and a confidence threshold of $0.999$, and set the BANSAC stopping criteria threshold $\tau$ to $0.01$ in BANSAC and $0.1$ in P-BANSAC (same parameters as in the main paper's results). Results are shown in \cref{tab:photoTourism_results}.

The results obtained are similar to those obtained in estimating the essential matrix. BANSAC is the best method in accuracy, followed by P-BANSAC in most scenes. Both are also the fastest methods overall.

\subsection{Synthetic data}\label{supp:synthetic}

We consider two simple problems: curve and circle-fitting. For each, we ran several experiments, varying the inlier rate between $15$ and $50\%$. Each experiment has 300 data points ranging between $\left[-1,1\right]$. Inliers are disturbed by a Gaussian noise of mean $0$ and variance $0.02$, and outliers are modeled by a uniform distribution with a maximum absolute value of $1.0$. We evaluate BANSAC against RANSAC and BaySAC, which we implemented from scratch since no code is available. As estimation parameters, we use an error threshold value of $0.02$, $3000$ maximum iterations, and an estimation confidence of $0.99$. In BANSAC, the initial probabilities $\mathcal{P}^0$ are set to $0.5$ for all data points, and the stopping criterion threshold $\tau$ is set to $0.01$. We measure the root mean squared error (RMSE) of the geometric distance of points in the estimated model to the desired model and the number of iterations made. We present the mean values obtained after $1000$ randomly generated trials. The results are shown in \cref{fig:results_synthetic_data}.

\begin{figure}
    \begin{subfigure}[t]{0.49\linewidth}
        \includegraphics[width=1\linewidth]{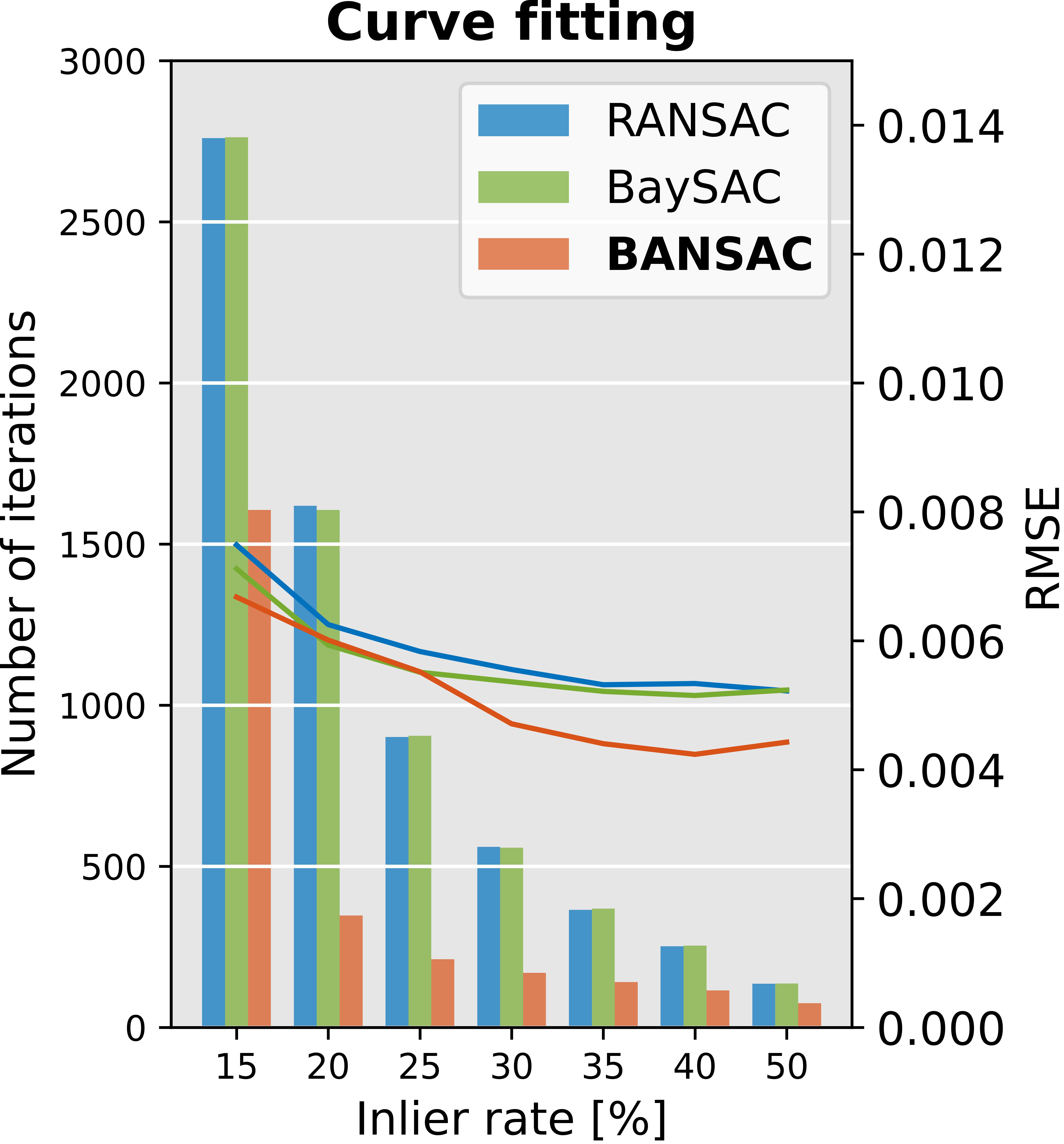}
    \end{subfigure}
    \begin{subfigure}[t]{0.49\linewidth}
        \includegraphics[width=1\linewidth]{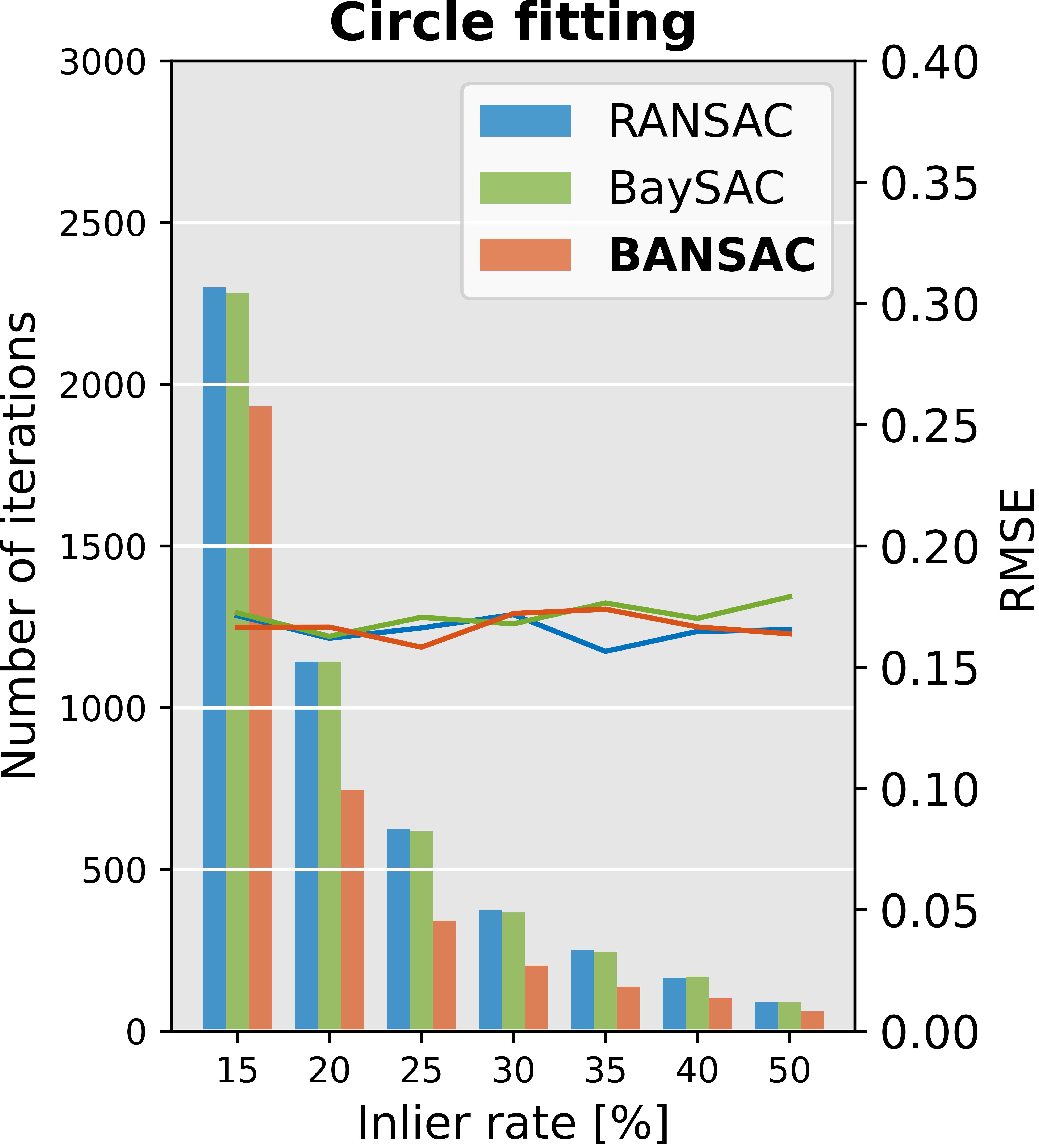}
    \end{subfigure}
    \caption{\it Experimental results for the curve (left) and circle (right) fitting. We compare RANSAC, BaySAC, and BANSAC based on the number of iterations and RMS error for different inlier rates.}
    \label{fig:results_synthetic_data}
\end{figure}

We observe that BANSAC has an accuracy similar to or better than the baselines requiring significantly fewer iterations, even for lower inlier rates. \Cref{fig:curve_fitting_example} illustrates the BANSAC probability update for the curve fitting problem.

\begin{figure}
        \includegraphics[width=1\linewidth]{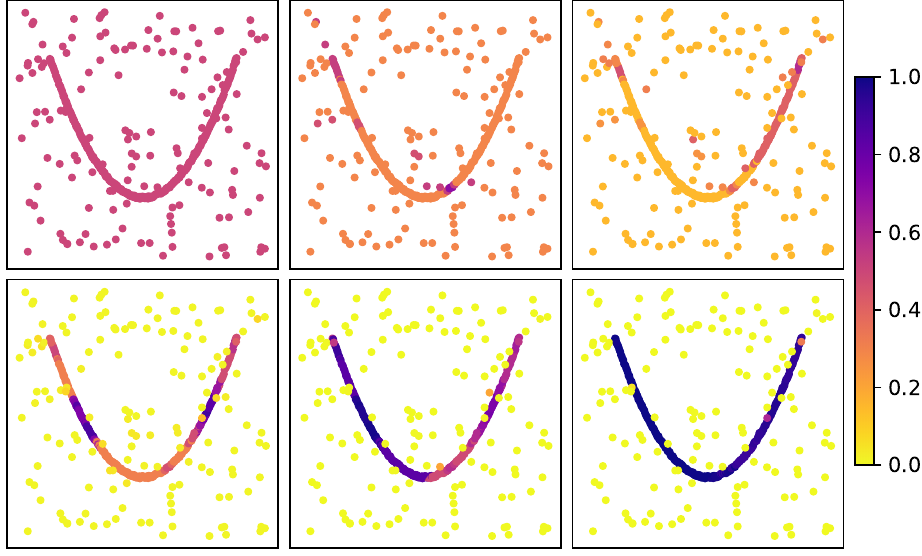}%
        \caption{\it Example of BANSAC inlier probability update over iterations for a curve fitting problem (color code at the right). In the first row, from left to right, we show iterations $0$, $2$, and $4$. Second row shows iteration $9$, $11$, and $14$.}
        \label{fig:curve_fitting_example}
\end{figure}

\subsection{Ablation studies}\label{supp:abaltion}

Next, we test different configurations for three components of the proposed algorithm. We present experiments using diverse conditional probability tables (CPTs) parameters, various activation functions for sampling, and combinations of different stopping criteria. The results were obtained using PhotoTourism sequence {\tt sacre\_coeur} (all pairs) for the uncalibrated relative pose problem (fundamental matrix estimation), using an error threshold of $0.5$, $10000$ maximum iterations, and a confidence of $0.999$, as in the main paper. BANSAC stopping criterion threshold $\tau$ is set to $0.01$. We evaluate the mAA of the rotation and translation errors at 5 and 10 degrees and the average execution time.

\subsubsection{Conditional probability tables}
To infer $P(x_n^k=\text{inlier}\ |\ c_n^{1:k})$ we need to define the CPTs of $P(c_n^k\ | \ x_n^{k-1})$ and $P(x_n^{k-1}\ | \ c_n^k, x_n^{k-1})$ for the 1st order Markov assumption. We present these CPTs in \cref{tab:prob_C}.
\begin{table}[t]
    \caption{\it Conditional probability table of $P(c_{k}^{n}|\mathbf{x}_{k-1}^{n})$ and $P(\mathbf{x}_{k}^{n}|c_{k}^{n},\mathbf{x}_{k-1}^{n})$.}
    \label{tab:prob_C}
    \begin{minipage}{0.205\textwidth}
        \centering
        \resizebox{1\linewidth}{!}{\setlength{\tabcolsep}{3.5pt}\begin{tabular}{cccc}
            \toprule
            $c_n^{k}$ & $x_n^{k-1}$ & $P(c_n^{k}\ | \ x_n^{k-1})$ \\\midrule
            Inlier & Inlier & $\gamma(\epsilon^k)$ \\
            Inlier & Outlier & $1 - \gamma(\epsilon^k)$ \\
            \bottomrule
        \end{tabular}
        }
    \end{minipage} \hfill
    \begin{minipage}{0.25\textwidth}
        \centering    
        \resizebox{1\linewidth}{!}{\setlength{\tabcolsep}{1.5pt}\begin{tabular}{ccccc}
            \toprule
            $x_n^{k}$ & $x_n^{k-1}$ & $c_n^{k}$ & $P(x_n^{k}\ |\ c_n^{k},x_n^{k-1})$ \\\midrule
            Inlier & Inlier & Inlier & $1.0$ \\
            Inlier & Inlier & Outlier & $1.0$ \\
            Inlier & Outlier & Inlier & $0.2$ \\
            Inlier & Outlier & Outlier & $0.0$ \\ 
            \bottomrule  
        \end{tabular}
        }
    \end{minipage}
\end{table}
The values for the CPT of $P(x_n^{k-1}\ | \ c_n^k, x_n^{k-1})$ were obtained empirically after testing different variations. We found that probability update is robust to slight variations of the reported parameters. The parameters of the CPT of $P(c_n^k\ | \ x_n^{k-1})$ are defined using a function $\gamma(\cdot)$. We want this function to give a high probability to classifications made by good models and vice versa. Since the quality of a model is defined by its inlier ratio, we define this function as $\gamma(\epsilon^k)$, where $\epsilon^k$ is the inlier ratio at iteration $k$. We test the following functions $\gamma(\epsilon^k)$ (variations of these functions with different values were tested, we are listing the ones that produced the best results):
\begin{align}
    \gamma_1(\epsilon^k) &=
    \begin{cases}
        0.62 \cdot \epsilon^k + 0.5, & \epsilon^k < 0.7143 \\
        0.2 \cdot \epsilon^k + 0.8, & \text{otherwise}
    \end{cases},
    \\
    \gamma_2(\epsilon^k) &= \frac{0.5}{0.5+e^{-10 \cdot (\epsilon^k - 0.3)}}, \text{and}
    \\
    \gamma_3(\epsilon^k) &= \text{tanh}(3 \cdot \epsilon^k).
\end{align}
We present results using these functions with BANSAC and P-BANSAC in \cref{table:ablation_classification_CPT}.
\begin{table}[t]
    \centering
    \caption{\it Evaluation of BANSAC using different activation functions to define the parameters of the conditional probability table of $P(c_n^{k}\ |\ x_n^{k-1})$.}
    \label{table:ablation_classification_CPT}
    \resizebox{1\linewidth}{!}{%
        \setlength{\tabcolsep}{13.5pt}\begin{NiceTabular}{@{}lccc}[code-before =%
        \rectanglecolor{Gray!20}{1-2}{14-4}%
        ]
        \toprule
        \multirow{2}{*}{\makecell{\thead{Metrics}}} & \multicolumn{3}{c}{\makecell{\thead{Activation functions}}} \\ \cmidrule(lr){2-4}
        & \makecell{$\gamma_1(\psi)$} & \makecell{$\gamma_2(\psi)$} & \makecell{$\gamma_3(\psi)$} \\ \midrule
        Rotation mAA $(5^\circ)$ \ $\uparrow$      & 0.836 & 0.793 & 0.790 \\
        Rotation mAA $(10^\circ)$ \ $\uparrow$     & 0.864 & 0.827 & 0.827 \\
        Translation mAA $(5^\circ)$ \ $\uparrow$   & 0.775 & 0.738 & 0.732 \\
        Translation mAA $(10^\circ)$ \ $\uparrow$  & 0.825 & 0.795 & 0.791 \\
        Avg. execution time $\left[ ms \right]$ \ $\downarrow$ & 13.9 & 17.4 & 17.5 \\
        \bottomrule
    \end{NiceTabular}
    }%
\end{table}

We achieved the best results in accuracy and execution time when using $\gamma_1(\psi)$. Based on these experiments, we decided to use $\gamma_1(\psi)$ in $P(c_n^{k}\ |\ x_n^{k-1})$ in all the experiments.

In the experiment shown in the main paper where we use the 2nd and 3rd orders of the Markov assumption, we use the CPTs shown in \cref{tab:cpt_2nd_markov,tab:cpt_3rd_markov}, respectively.
\begin{table}[t]
    \caption{\it Conditional probability table of $P(x_n^{k}\ |\ c_n^{k},x_n^{k-2:k-1})$.}
    \label{tab:cpt_2nd_markov}
    \centering    
    \resizebox{1\linewidth}{!}{\setlength{\tabcolsep}{9.0pt}\begin{tabular}{cccccc}
        \toprule
        $x_n^{k}$ & $x_n^{k-1}$ & $x_n^{k-2}$ & $c_n^{k}$ & $P(x_n^{k}\ |\ c_n^{k},x_n^{k-2:k-1})$ \\\midrule
        Inlier & Inlier  & Inlier  & Inlier  & $1.0$ \\
        Inlier & Inlier  & Inlier  & Outlier & $0.8$ \\
        Inlier & Inlier  & Outlier & Inlier  & $0.9$ \\
        Inlier & Inlier  & Outlier & Outlier & $0.7$ \\ 
        Inlier & Outlier & Inlier  & Inlier  & $0.2$ \\
        Inlier & Outlier & Inlier  & Outlier & $0.1$ \\
        Inlier & Outlier & Outlier & Inlier  & $0.1$ \\
        Inlier & Outlier & Outlier & Outlier & $0.0$ \\ 
        \bottomrule  
    \end{tabular}
    }
\end{table}
\begin{table}[t]
    \caption{\it Conditional probability table of $P(x_n^{k}\ |\ c_n^{k},x_n^{k-3:k-1})$.}
    \label{tab:cpt_3rd_markov}
    \centering    
    \resizebox{1\linewidth}{!}{\setlength{\tabcolsep}{5.0pt}\begin{tabular}{ccccccc}
        \toprule
        $x_n^{k}$ & $x_n^{k-1}$ & $x_n^{k-2}$ & $x_n^{k-3}$ & $c_n^{k}$ & $P(x_n^{k}\ |\ c_n^{k},x_n^{k-3:k-1})$ \\\midrule
        Inlier & Inlier  & Inlier  & Inlier  & Inlier  & $1.0$  \\
        Inlier & Inlier  & Inlier  & Inlier  & Outlier & $0.8$  \\
        Inlier & Inlier  & Inlier  & Outlier & Inlier  & $0.9$  \\
        Inlier & Inlier  & Inlier  & Outlier & Outlier & $0.7$  \\ 
        Inlier & Inlier  & Outlier & Inlier  & Inlier  & $0.6$  \\
        Inlier & Inlier  & Outlier & Inlier  & Outlier & $0.5$  \\
        Inlier & Inlier  & Outlier & Outlier & Inlier  & $0.4$  \\
        Inlier & Inlier  & Outlier & Outlier & Outlier & $0.2$  \\ 
        Inlier & Outlier & Inlier  & Inlier  & Inlier  & $0.3$  \\
        Inlier & Outlier & Inlier  & Inlier  & Outlier & $0.2$  \\
        Inlier & Outlier & Inlier  & Outlier & Inlier  & $0.1$  \\
        Inlier & Outlier & Inlier  & Outlier & Outlier & $0.3$  \\ 
        Inlier & Outlier & Outlier & Inlier  & Inlier  & $0.2$  \\
        Inlier & Outlier & Outlier & Inlier  & Outlier & $0.1$  \\
        Inlier & Outlier & Outlier & Outlier & Inlier  & $0.05$ \\
        Inlier & Outlier & Outlier & Outlier & Outlier & $0.0$  \\ 
        \bottomrule  
    \end{tabular}
    }
\end{table}
Similar to the CPT for the 1st order of the Markov assumption, the outlined parameters were obtained empirically.

\subsubsection{Weighted sampling}

In each iteration $k$, we perform a sampling weighted by the probabilities estimated in the previous iteration $\mathcal{P}^{k-1}$. Instead of simply using the probability values directly, we test the use of activation functions to increase the range of weights. The goal is to increase the chances of choosing points with higher inlier probabilities. We tested the following activation functions (different functions were tested, and we are showing the ones that gave the best results):
\begin{align}
    \rho_1(\psi) &= \psi \cdot 100
    \\
    \rho_2(\psi) &= \begin{cases}
                        100\cdot\psi & \psi > 0.3 \\
                        10\cdot\psi & \text{otherwise}
                    \end{cases},
    \\
    \rho_3(\psi) &= \frac{100}{1+e^{-10 \cdot (\psi - 0.5)}}
    \\
    \rho_4(\psi) &= 130 \cdot \text{tanh}(\psi)
\end{align}
where $\psi \triangleq P(x_n^k=\text{inlier}\ |\ C_n^{1:k})$ is the estimated probability for the $\text{n}^{\text{th}}$ data point at iteration $k$. In \cref{table:ablation_sampling}, we show results using these activations functions in BANSAC.
\begin{table}[t]
    \centering
    \caption{\it Evaluation of BANSAC using different activation functions to generate sampling weights.}
    \label{table:ablation_sampling}
    \resizebox{1\linewidth}{!}{
    \setlength{\tabcolsep}{10.7pt}\begin{NiceTabular}{@{}lcccc}[code-before =%
        \rectanglecolor{Gray!20}{1-2}{14-5}%
        ]
        \toprule
        \multirow{2}{*}{\thead{Metrics}} & \multicolumn{4}{c}{\thead{Sampling activation functions}} \\ \cmidrule(lr){2-5}
        & \makecell{$\rho_1(\psi)$} & \makecell{$\rho_2(\psi)$} & \makecell{$\rho_3(\psi)$} & \makecell{$\rho_4(\psi)$} \\ \midrule
            Rotation mAA $(5^\circ)$ \ $\uparrow$                  & 0.836 & 0.825 & 0.823 & 0.834 \\
            Rotation mAA $(10^\circ)$ \ $\uparrow$                 & 0.864 & 0.853 & 0.851 & 0.861 \\
            Translation mAA $(5^\circ)$ \ $\uparrow$               & 0.775 & 0.760 & 0.758 & 0.776 \\
            Translation mAA $(10^\circ)$ \ $\uparrow$              & 0.825 & 0.813 & 0.811 & 0.826 \\
            Avg. execution time $\left[ ms \right]$ \ $\downarrow$ & 13.9  & 13.5  & 13.6  & 14.2  \\
        \bottomrule
    \end{NiceTabular}
    }%
\end{table}

Of the tested functions, only $\rho_1(\psi)$ is linear. This function equally converts all points probabilities to the desired sampling range. The remaining give greater weights to points with higher probabilities and vice versa. Overall, we observed that $\rho_1(\psi)$ was the one that gave better results in accuracy and execution time. Based on these experiments, we chose to use $\rho_1(\psi)$ in all other experiments.

\subsubsection{Stopping criteria}
Finally, we assess the different kinds and combinations of stopping criteria we can use with our method: RANSAC, SPRT, PROSAC, BANSAC, and BANSAC combined with RANSAC, SPRT, or PROSAC. We show results using these different combinations of stopping criteria in \cref{tab:stopping_criteria}.

\begin{table}[t]
    \caption{\it Evaluation of BANSAC with different stopping criteria.}
    \label{tab:stopping_criteria}
    \resizebox{1\linewidth}{!}{
    \setlength{\tabcolsep}{1.7pt}\begin{NiceTabular}{@{}ccccccccc}[code-before =%
        \rectanglecolor{Gray!20}{1-5}{19-9}%
        ]
        \toprule
        \multicolumn{4}{c}{\thead{Stopping Criteria}} & \multicolumn{5}{c}{\thead{Results}} \\ \cmidrule(lr){1-4} \cmidrule(lr){5-9}
        \multirow{2}{*}{\makecell{RANSAC}} & \multirow{2}{*}{\makecell{SPRT}} & \multirow{2}{*}{\makecell{PROSAC}} & \multirow{2}{*}{\makecell{BANSAC}}  &
        \multicolumn{2}{c}{\makecell{Rotation}} & \multicolumn{2}{c}{\makecell{Translation}} & \makecell{Time} \\
        & & & & \makecell{mAA$(5^\circ)$} & \makecell{mAA$(10^\circ)$} & \makecell{mAA$(5^\circ)$} & \makecell{mAA$(10^\circ)$} & \makecell{Avg. $\left[ ms \right]$} \\ \midrule
        {\large\checkmark} & & &                    & 0.845 & 0.868 & 0.792 & 0.837 & 16.2 \\
        & {\large\checkmark} & &                    & 0.837 & 0.864 & 0.775 & 0.825 & 13.9 \\
        & & {\large\checkmark} &                    & 0.839 & 0.865 & 0.782 & 0.829 & 15.1 \\
        & & & {\large\checkmark}                    & 0.850 & 0.870 & 0.818 & 0.854 & 33.5 \\
        {\large\checkmark} & & & {\large\checkmark} & 0.845 & 0.867 & 0.793 & 0.837 & 16.1 \\
        & {\large\checkmark} & & {\large\checkmark} & 0.836 & 0.864 & 0.775 & 0.825 & 13.9 \\
        & & {\large\checkmark} & {\large\checkmark} & 0.838 & 0.864 & 0.782 & 0.829 & 14.4 \\
        \bottomrule
    \end{NiceTabular}}
\end{table}

We observe that, although BANSAC stopping criteria ensure the output results are accurate, it is slow. However, when we combine our stopping condition with others, we consistently improve execution time with a slight drop in accuracy.

\section{Other Markov Assumptions}\label{sec:other_markov_assumptions}

\begin{algorithm*}[t]
    \scriptsize
    \caption{ {\it BANSAC algorithm outline. In the algorithm below, $I$ means inlier and $O$ outlier.}\newline
     {\bf Input} -- Data $\mathcal{Q}$, and without pre-computed scores \newline
     {\bf Output} -- Best model $\theta^*$, and $\mathcal{C}^*$
    }\label{alg:pseudo-code}
    $k\gets 1$; \\
    $\Phi^+_n \gets 0.5,\ \forall n$ \Comment*{for $x_n^k = \text{true}$ (a pre-computed score can be used here)}
    $\Phi^-_n \gets 0.5,\ \forall n$ \Comment*{for $x_n^k = \text{false}$ (a pre-computed score can be used here)}
    $P_n = \frac{\Phi^+_n}{\Phi^+_n + \Phi^-_n}$ \Comment*{current weights used for sampling}
    \While{$k < K$}{
        ...\\
        Other RANSAC steps as listed in the main paper\;
        ...\\
        \For{all $n$}{
            \eIf{$c_n^k = I$}{
                $\widehat{\Phi}^+_n \gets P(x_n^k=I,c_n^k=I,x_n^{k-1}=I)P(c_n^k=I,x_n^{k-1}=I)\Phi^+_n + P(x_n^k=I,c_n^k=I,x_n^{k-1}=O)P(c_n^k=I,x_n^{k-1}=O)\Phi^-_n$\;
                $\widehat{\Phi}^-_n \gets P(x_n^k=O,c_n^k=I,x_n^{k-1}=I)P(c_n^k=I,x_n^{k-1}=I)\Phi^+_n + P(x_n^k=O,c_n^k=I,x_n^{k-1}=O)P(c_n^k=I,x_n^{k-1}=O)\Phi^-_n$\;
            }
            {
                $\widehat{\Phi}^+_n \gets P(x_n^k=I,c_n^k=O,x_n^{k-1}=I)P(c_n^k=O,x_n^{k-1}=I)\Phi^+_n + P(x_n^k=I,c_n^k=O,x_n^{k-1}=O)P(c_n^k=O,x_n^{k-1}=O)\Phi^-_n$\;
                $\widehat{\Phi}^-_n \gets P(x_n^k=O,c_n^k=O,x_n^{k-1}=I)P(c_n^k=O,x_n^{k-1}=I)\Phi^+_n + P(x_n^k=O,c_n^k=O,x_n^{k-1}=O)P(c_n^k=O,x_n^{k-1}=O)\Phi^-_n$\;
            }
            $\Phi^+_n \gets \widehat{\Phi}^+_n$\;
            $\Phi^-_n \gets \widehat{\Phi}^-_n$\;
            $P_n = \frac{\Phi^+_n}{\Phi^+_n + \Phi^-_n}$\;
        }
        ...\\
        Other RANSAC steps as listed in the main paper\;
        ...\\
    }
\end{algorithm*}

In this section, we present new derivations on probability updates. We show how to get exact inferences for second and third-order Markov assumptions.

\subsection{Second-order Markov assumption}

For the second-order assumption, in addition to the conditional independence constraints presented in the main paper, we have
\begin{equation}
    x_n^{j} \perp x_n^{0:j-3}\ |\ x_{n}^{j-1}, x_{n}^{j-2}, c_{n}^{j}  \ \ \forall j,
\end{equation}
which means
\begin{equation}
    P(x_n^j \  |\ x_n^{0:j-1}, c_n^{j}) = P(x_n^j \  |\ x_n^{j-2:j-1}, c_n^{j}), \ \ \forall j.
\end{equation}
Now, similar to what is done in the main document, by applying the chain rule of probabilities, we write the joint probability at iteration $k$ as
\begin{equation}
    \widetilde{P}(x_n^{0:k},c_n^{1:k}) = P(x_n^0) \prod_{j=1}^{k} \widetilde{\phi}(x_n^{j}, c_n^{j}) ,
    \label{eq:joint_prob_2}
\end{equation} 
where
{\small
\begin{equation}
    \widetilde{\phi}(x_n^{j}, c_n^{j}) =
    \begin{cases}
    P(x_n^j \  |\ x_n^{j-2:j-1}, c_n^{j}) P(c_n^{j}\ |\ x_n^{j-1}), & j \geq 2 \\
    P(x_n^1 \  |\ x_n^{0}, c_n^{1}) P(c_n^{1}\ |\ x_n^{0}), & j = 1
    \end{cases}.
    \label{eq:phi_markov_blanket_2}
\end{equation}}%
We use $\widetilde{P}(.)$ to distinguish from the joint probability derived in the main document.
Following the same steps derived in the main document, from \cref{eq:joint_prob_2,eq:phi_markov_blanket_2} the exact inference is given by
\begin{equation}
    P(x_n^k=\text{inlier}\ |\ C_n^{1:k}) = \alpha \widetilde{\Phi}(x_n^{k}=\text{inlier}, x_n^{0:k-1}, C_n^{1:k}),
    \label{eq:inference_2}
\end{equation}
where again $\alpha$ is the normalization factor, and 
\begin{multline}
\widetilde{\Phi}(x_n^k, x_n^{0:k-1},C_n^{1:k}) = \sum_{x_n^{k-1}} \widetilde{\Phi}^{\dag}(x_n^k, x_n^{0:k-1},C_n^{1:k})
\end{multline}
where
\begin{multline}
\widetilde{\Phi}^\dag(x_n^k, x_n^{0:k-1},C_n^{1:k}) = \\
\sum_{x_n^{k-2}} \widetilde{\phi}(x_n^{k},C_n^{k})
\sum_{x_n^{k-3}} \widetilde{\phi}(x_n^{k-1},C_n^{k-1}) \\
\cdots
\sum_{x_n^1} \widetilde{\phi}(x_n^{3},C_n^{3})
\sum_{x_n^0} \widetilde{\phi}(x_n^{2},C_n^{2}) \widetilde{\phi}(x_n^{1},C_n^{1}) P(x_n^0).
\label{eq:Phi_2}
\end{multline}
As in the main document, a convenient result of \cref{eq:Phi_2} is that $\widetilde{\Phi}^{\dag}(.)$ can be calculated recursively as follows:
\begin{multline}
    \widetilde{\Phi}^{\dag}(x_n^k, x_n^{0:k-1}, C_n^{1:k}) = \\
    \begin{cases}
    \sum_{x_n^{k-2}} \widetilde{\phi}(x_n^{k}, C_n^{k}) \widetilde{\Phi}^{\dag}(x_n^{k-1},x_n^{0:k-2},C_n^{1:k-1}) & k \geq 2 \\
    \widetilde{\phi}(x_n^{1},C_n^{1}) P(x_n^0) & k = 1 
    \end{cases}.
\end{multline}

For the second-order Markov assumption experiments, the only difference compared to what is described for the first-order is the use of the conditional probability in \cref{eq:inference_2} as the sampling weights.

\subsection{Third-order Markov assumption}

For the third-order Markov assumption, we have the conditional independence constraints
\begin{equation}
    x_n^{j} \perp x_n^{0:j-4}\ |\ x_{n}^{j-1}, x_{n}^{j-2}, x_{n}^{j-3}, c_{n}^{j} \ \ \forall j,
\end{equation}
which means
\begin{equation}
    P(x_n^j \  |\ x_n^{0:j-1}, c_n^{j}) = P(x_n^j \  |\ x_n^{j-3:j-1}, c_n^{j}), \ \ \forall j.
\end{equation}
Again, by applying the chain rule of probabilities, we write the joint probability at iteration $k$ as 
\begin{equation}
    \widetilde{\widetilde{P}}(x_n^{0:k},c_n^{1:k}) = P(x_n^0) \prod_{j=1}^{k} \widetilde{\widetilde{\phi}}(x_n^{j}, c_n^{j}) ,
    \label{eq:joint_prob_3}
\end{equation} 
where
{\small
\begin{equation}
    \widetilde{\widetilde{\phi}}(x_n^{j}, c_n^{j}) =
    \begin{cases}
    P(x_n^i \  |\ x_n^{j-3:j-1}, c_n^{j}) P(c_n^{j}\ |\ x_n^{j-1}), & j \geq 3 \\
    P(x_n^2 \  |\ x_n^{0:1}, c_n^{2}) P(c_n^{2}\ |\ x_n^{1}), & j = 2  \\
    P(x_n^1 \  |\ x_n^{0}, c_n^{1}) P(c_n^{1}\ |\ x_n^{0}), & j = 1
    \end{cases}.
    \label{eq:phi_markov_blanket_3}
\end{equation}}%
Following the same steps shown in the main document, from \cref{eq:joint_prob_3,eq:phi_markov_blanket_3} the exact inference is given by
\begin{equation}
    P(x_n^k=\text{inlier}\ |\ C_n^{1:k}) = \alpha \widetilde{\widetilde{\Phi}}(x_n^{k}=\text{inlier}, x_n^{0:k-1}, C_n^{1:k}),
    \label{eq:inference_3}
\end{equation}
where again $\alpha$ is the normalization factor, and 
\begin{multline}
\widetilde{\widetilde{\Phi}}(x_n^k, x_n^{0:k-1},C_n^{1:k}) = \sum_{x_n^{k-1}} \sum_{x_n^{k-2}} \widetilde{\widetilde{\Phi}}^{\dag}(x_n^k, x_n^{0:k-1},C_n^{1:k}),
\end{multline}
where
\begin{multline}
\widetilde{\widetilde{\Phi}}^\dag(x_n^k, x_n^{0:k-1},C_n^{1:k}) = \\
\sum_{x_n^{k-3}} \widetilde{\widetilde{\phi}}(x_n^{k},C_n^{k})
\sum_{x_n^{k-4 }} \widetilde{\widetilde{\phi}}(x_n^{k-1},C_n^{k-1}) \cdots \\
\sum_{x_n^1} \widetilde{\widetilde{\phi}}(x_n^{4},C_n^{4}) \sum_{x_n^0}\widetilde{\widetilde{\phi}}(x_n^{3},C_n^{3}) \widetilde{\widetilde{\phi}}(x_n^{2},C_n^{2}) \widetilde{\widetilde{\phi}}(x_n^{1},C_n^{1}) P(x_n^0).
\label{eq:Phi_3}
\end{multline}
Again, we can write \cref{eq:Phi_3} in a recursive way:
\begin{multline}
    \widetilde{\widetilde{\Phi}}^{\dag}(x_n^k, x_n^{0:k-1}, C_n^{1:k}) = \\
    \begin{cases}
    \sum_{x_n^{k-3}} \widetilde{\widetilde{\phi}}(x_n^{k}, C_n^{k}) \widetilde{\widetilde{\Phi}}^{\dag}(x_n^{k-1},x_n^{0:k-2},C_n^{1:k-1}) & k \geq 3 \\
    \widetilde{\widetilde{\phi}}(x_n^{2},C_n^{2}) \widetilde{\widetilde{\phi}}(x_n^{1},C_n^{1}) P(x_n^0) & k = 2 \\
    \widetilde{\widetilde{\phi}}(x_n^{1},C_n^{1}) P(x_n^0) & k = 1 
    \end{cases}.
\end{multline}
Note for $k = 1$, \cref{eq:Phi_3} does not sum in $x_n^{-2}$, since there is no such variable.

Finally, the weights for the sampling are taken from the inference in \cref{eq:inference_3}.

\subsection{Probability Updates Pseudo-code}
\label{app:pseudo-code}

The probability updates derived in this code are easy to implement. An algorithm with the pseudo-code for the first-order Markov assumption is shown in \cref{alg:pseudo-code}, in which probabilities are taken from \cref{tab:prob_C}. The second and third-order constraints are derived similarly.

}{}

\end{document}